\documentclass[sigconf,nonacm,table,dvipsnames]{acmart}

\usepackage{listings}
\usepackage{booktabs}
\usepackage{array}
\usepackage{multirow}
\usepackage{enumitem}
\usepackage{lineno}
\usepackage{subcaption}
\usepackage{tikz}
\usepackage{tabularx}

\usepackage{algorithm}
\usepackage{algpseudocode}

\usetikzlibrary{
  positioning,
  arrows.meta,
  shapes.geometric,
  fit,
  backgrounds,
  calc
}

\DeclareUnicodeCharacter{2264}{\ensuremath{\leq}}

\renewcommand\footnotetextcopyrightpermission[1]{} 

\AtBeginDocument{%
  \providecommand\BibTeX{{%
    \normalfont B\kern-0.5em{\scshape i\kern-0.25em b}\kern-0.8em\TeX}}}

\graphicspath{{./images/}} 
\begin{document}

\title{Catastrophic Forgetting in Sequential Thermal Anti-UAV Detection}
\subtitle{The Role of Scale-Conditioned Gradient Imbalance}

\author{Khac Duc Giang Nguyen}
\affiliation{%
  \institution{University of Amsterdam}
  \city{Amsterdam}
  \country{The Netherlands}
}
\email{giang.nguyen5@student.uva.nl}

\author{Seyed Sahand Mohammadi Ziabari}
\authornote{Corresponding author.}
\affiliation{%
  \institution{University of Amsterdam}
  \city{Amsterdam}
  \country{The Netherlands}
}
\affiliation{%
  \institution{Department of Computer Science and Technology, SUNY Empire State University}
  \city{Saratoga Springs}
  \state{NY}
  \country{USA}
}
\email{s.s.mohammadiziabari@uva.nl}

\author{Ali Mohammed Mansoor Alsahag}
\affiliation{%
  \institution{University of Amsterdam}
  \city{Amsterdam}
  \country{The Netherlands}
}
\email{a.m.m.alsahag@uva.nl}

\begin{abstract}

{
Counter-UAV systems based on thermal infrared detection must stay accurate
as operational datasets evolve, yet sequential fine-tuning causes
catastrophic forgetting of prior tasks, a problem not previously
quantified in this domain. This continual-learning study measures the
stability-plasticity trade-off in YOLOMG, a YOLOv5-based detector run as a
single thermal-infrared stream (motion channel disabled), trained sequentially
across three anti-UAV benchmarks of rising scale difficulty: Anti-UAV-RGBT,
Anti-UAV410, and CST Anti-UAV. Naive fine-tuning on CST yields a Forgetting
Measure of $-0.605$ against the Stage~1 ceiling (90\% capability loss),
$-0.572$ of it in Stage~3 alone, whereas knowledge distillation from a
frozen teacher is associated with $\mathrm{FM} = -0.033 \pm 0.004$ across
three seeds (95\% retention); because no Stage~2 no-KD control is included,
this result establishes retention under KD training rather than a causal KD effect. Per-stratum analysis shows large-target
detection collapsing to near-zero within the first epoch, despite an
inter-stage cosine similarity of 0.987 over the gradient-updated
weights, pointing to a scale-conditioned gradient imbalance, rather than
weight drift, as a candidate mechanism. Scale-Stratified Herding (SSH), a
300-exemplar buffer balanced across four UAV size strata, roughly halves
the forgetting ($\mathrm{FM} = -0.605 \to -0.311$) and keeps large-target
detection non-zero; an ablation attributes the gain to the scale
stratification rather than the herding: random-stratified replay performs
at least as well ($\mathrm{FM}=-0.221$ versus $-0.311$ for SSH). These replay
results are single-seed and should be treated as preliminary.
}

\end{abstract}

\keywords{Continual Learning, Scale-Distribution Shift, Catastrophic Forgetting, Thermal Anti-UAV Detection, Knowledge Distillation, Gradient Starvation} 


\fancyhead{}
\maketitle
\section{Introduction}
\label{sec:introduction}

The proliferation of commercial drones has created urgent demand for
counter-unmanned-aerial-vehicle (C-UAV) surveillance systems capable of
reliable detection across diverse operational
environments~\cite{zhao_vision-based_2022}. Thermal infrared (TIR)
imaging is increasingly favoured over visible-spectrum cameras for this
role: it operates independently of ambient lighting, remains effective
against low-visibility backgrounds, and provides a distinct radiometric
signature for airborne targets~\cite{bohuang_410,zhu_evidential_2023}.
Deep learning detectors, particularly YOLO-family
architectures~\cite{jocher2020yolov5}, perform strongly on individual TIR
anti-UAV benchmarks~\cite{guo_yolomg_2025}. But practical C-UAV systems
must stay accurate as new datasets emerge, sensors are upgraded, and the
threat profile shifts toward smaller, harder-to-detect targets.

The core technical challenge this creates is catastrophic
forgetting~\cite{mccloskey1989catastrophic,goodfellow2013empirical}: when
a neural network is sequentially fine-tuned on a new dataset, the weight
updates that improve performance on the new task interfere with
representations learned for previous ones, causing abrupt performance
degradation on earlier
tasks~\cite{mermillod2013stability,Kirkpatrick_2017}. The field of
continual learning (CL) exists to manage this
stability-plasticity trade-off~\cite{wang2024comprehensive}. Rehearsal
methods store a subset of past training data and interleave it during
subsequent training~\cite{rebuffi2017icarl,chaudhry2019tiny,deng_unlocking_2025};
regularisation methods penalise changes to important
parameters~\cite{Kirkpatrick_2017,lopezpaz2022gradientepisodicmemorycontinual};
gradient projection methods constrain new-task gradient updates to
subspaces orthogonal to previous tasks'
gradients~\cite{farajtabar2019ogd,saha2021gpm}; and knowledge distillation (KD) methods
use a frozen teacher model to constrain the updated model's output
distribution~\cite{hinton2015distilling,li2017lwf}.
Shmelkov et al.~\cite{shmelkov2017iod} introduced distillation-based CL
for general object detection, showing that distillation can reduce forgetting
when new classes are added. To our knowledge, these approaches have not been
systematically evaluated in the thermal anti-UAV domain, where distributional
shifts may include sensor, scene, weather, and target-scale changes.

Existing thermal anti-UAV benchmarks capture complementary facets of
this problem in isolation. Anti-UAV-RGBT~\cite{jiang_anti-uav_2023}
provides 318 paired RGB-TIR video streams of standard-scale drones;
Anti-UAV410~\cite{bohuang_410} contains 410 purely thermal sequences
spanning a wide range of drone sizes; and CST
Anti-UAV~\cite{xie2025cstantiuavthermalinfrared} focuses on extreme tiny
targets in cluttered thermal scenes, with 87.9\% tiny and 11.9\% small
bounding boxes and effectively no large targets. We are not aware of prior work that has chained these
datasets into a single model's training history and measured the
resulting stability-plasticity trade-off. Prior continual learning work
has also not examined gradient starvation~\cite{pezeshki2021gradient}, a
phenomenon in which dominant features suppress gradient signals for
weaker but relevant ones, as a driver of forgetting after distributional
shift. This is the research gap this work addresses.

We investigate the following hypothesis and three research questions
using YOLOMG~\cite{guo_yolomg_2025}, a dual-modal YOLOv5-based detector
(run here with its motion channel disabled), trained sequentially on
Anti-UAV-RGBT (Stage 1), Anti-UAV410 (Stage 2), and CST Anti-UAV
(Stage 3). All three datasets are thermal infrared, so the input modality
is held constant across the curriculum; the shifts under study are the
moderate cross-dataset domain shift of Stage 2 and the extreme
scale-distribution shift of Stage 3.

H1: Under the training protocol used here, the Stage~3 transition to the
extreme tiny-target CST Anti-UAV distribution produces substantially greater
Task~1 forgetting than the Stage~2 transition. Because Stage~2 uses KD while
the Stage~3 baseline uses naive fine-tuning, this comparison is descriptive
rather than a causal isolation of distribution-shift type.

RQ1: Does knowledge distillation preserve Task 1 performance during
Task 2 training under domain shift? Specifically, does the KD loss term
in Stage 2 (Anti-UAV410) training prevent the model from forgetting its
Anti-UAV-RGBT detection capability, despite the thermal cross-dataset
domain shift between the two benchmarks?

RQ2: How does scale shift characterise the pattern of catastrophic
forgetting during Task 3 training? Specifically, does the extreme size
distribution of CST Anti-UAV (87.9\% tiny, 11.9\% small, 0\% large)
produce stratum-specific forgetting in which large-target
detection collapses disproportionately fast, and how is this reflected in
per-stratum mean Average Precision (mAP), the Forgetting Measure, parameter
drift, and the gradient signal reaching each detection head?

RQ3: Does re-injecting scale-stratified exemplars via replay mitigate the
Stage 3 forgetting? A 300-exemplar buffer that selects exemplars across four
UAV size strata (tiny, small, normal, large) by iCaRL-style greedy herding
restores the gradient signal for the under-represented strata during
Stage 3, and is compared against naive fine-tuning and a
random-stratified control that isolates the effect of the stratification.

In brief, sequential training produces substantial and measurable
forgetting. Stage~2 training with knowledge distillation retains most Task~1
performance, although the absence of a no-KD Stage~2 control prevents
attributing that retention causally to distillation. The Stage 3 results point to a scale-induced forgetting pattern
whose evidence is, tentatively, more consistent with a scale-conditioned
gradient imbalance than with outright weight overwriting. Scale-Stratified
Herding (SSH) roughly halves this forgetting, which suggests that
scale-stratified replay can offset the imbalance.

The remainder of this work is organised as follows.
Section~\ref{sec:related_work} surveys related work on continual
learning, thermal UAV detection, and knowledge distillation.
Section~\ref{sec:methodology} describes the experimental design, datasets,
and methods. Section~\ref{sec:results} presents the empirical results.
Section~\ref{sec:discussion} interprets the findings and discusses
limitations. Section~\ref{sec:conclusion} concludes with implications and
future directions.

\section{Related Work}
\label{sec:related_work}

To our knowledge, prior work has not combined a continual-learning protocol
with thermal anti-UAV detection across multiple real operational datasets or
examined scale-conditioned gradient imbalance, a form of gradient starvation,
as a candidate driver of forgetting under scale-distribution shift. This section situates the research within four
bodies of literature that together delimit this gap.

\subsection{Thermal Anti-UAV Detection}

Counter-UAV surveillance using thermal infrared imaging has received
growing attention due to the sensor's independence from ambient lighting
and its distinct radiometric signature for airborne targets. Zhao et
al.~\cite{zhao_vision-based_2022} provide a broad survey of vision-based
anti-UAV detection and tracking, identifying the shift from visible to
thermal modalities as a key trend driven by operational requirements at
night and in degraded visibility. Jiang et al.~\cite{jiang_anti-uav_2023}
introduced Anti-UAV-RGBT, a paired RGB-TIR benchmark of 318 video
sequences that remains the most widely used dual-modality anti-UAV
evaluation resource. Huang et al.~\cite{bohuang_410} extended this with
Anti-UAV410, a purely thermal benchmark of 410 sequences spanning a wider
range of acquisition conditions and drone distances. Both datasets focus
on standard-scale commercial quadcopters; the more recent CST
Anti-UAV~\cite{xie2025cstantiuavthermalinfrared} shifts attention to
extreme tiny targets in cluttered thermal scenes, creating the
near-complete scale inversion relative to Anti-UAV-RGBT that motivates
the forgetting analysis in this research. Detection in such conditions
shares challenges with the broader infrared small target detection (IRSTD)
literature, where methods such as DNA-Net~\cite{li2021dnanet} use dense
nested attention to preserve sub-16-pixel target features that deep
pooling layers would otherwise discard.

On the detector side, YOLO-family architectures~\cite{jocher2020yolov5}
have become a common choice for real-time anti-UAV detection because of their
favourable speed--accuracy balance. Guo et al.~\cite{guo_yolomg_2025}
extended YOLOv5 with a pixel-level motion-difference channel (YOLOMG), fusing
appearance and motion at three feature-pyramid scales (P3, P4, P5)~\cite{lin2017fpn}.
Recent anti-UAV work has also examined complementary reliability failures.
Suttorp et al.~\cite{suttorp2026uncertainty} studied uncertainty-aware
RGB--thermal fusion under modality disagreement, while Golovchits et
al.~\cite{golovchits2026evidential} evaluated evidential learning and
sensor-gating across thermal and multimodal anti-UAV settings. Birkental et
al.~\cite{birkental2026fog} quantified detector degradation under controlled
fog severity and showed that robustness can change sharply as visibility
deteriorates. Together, these studies reinforce that operational anti-UAV
reliability depends on distributional conditions beyond aggregate benchmark
accuracy, but they do not study catastrophic forgetting under sequential
cross-dataset adaptation. Zhu et al.~\cite{zhu_evidential_2023} likewise
investigated evidential detection and tracking, while domain adaptation from
RGB to thermal and across weather conditions has been explored by Zhang et
al.~\cite{zhang_ttsda-yolo_2025} and Zheng et al.~\cite{zheng_foggy_2025}.
These works address uncertainty or single-step robustness/adaptation rather
than retention of earlier datasets after subsequent training.

\subsection{Continual Learning}

The instability of neural networks under sequential training has been
studied since McCloskey and Cohen~\cite{mccloskey1989catastrophic}
documented catastrophic interference in connectionist models. Goodfellow
et al.~\cite{goodfellow2013empirical} provided the first systematic
empirical investigation of this phenomenon in modern gradient-based
networks, examining how activation functions, optimisers, and task
relationships modulate the degree of forgetting. Mermillod et
al.~\cite{mermillod2013stability} formalised the stability-plasticity
dilemma: a model must retain prior knowledge while acquiring new
capabilities, and optimising for one degrades the other.

Van de Ven et al.~\cite{vandeven2022three} formalise three CL scenarios:
task-incremental learning (task identity known at test time), domain-
incremental learning (same class set, changing input distribution), and
class-incremental learning (new classes arrive without task identity). Our
three-stage curriculum is domain-incremental: the target class (UAV) is
fixed across all stages and the input modality is held constant at thermal
infrared (only the IR stream is used), while the scale regime shifts from
standard-scale to wide-range to extreme tiny-target.

Continual learning is one of several responses to distribution shift.
Domain generalisation and one-shot domain adaptation aim to transfer a model
to a new distribution without necessarily preserving source-task
performance, while parameter-efficient adaptation---low-rank adapters (LoRA)
and related PEFT methods---has recently been applied to sequential and
domain-incremental settings, including object
detection~\cite{wistuba2023color,he2025cllora,li2025continualadaptation}.
This work adopts continual learning because the operational goal is
\emph{cumulative} retention across a fixed sequence of real anti-UAV
datasets, for which catastrophic forgetting is the defining failure
mode~\cite{wang2024comprehensive}. Replay is preferred over a low-rank
weight constraint because, as Section~\ref{sec:rq2} shows, the failure is a
missing \emph{scale-conditioned gradient}: replay re-injects that signal
directly, whereas constraining weight movement does not supply it.
Parameter-efficient continual adaptation nonetheless remains a promising
alternative and is revisited as future work.

The continual learning literature organises mitigation methods into four
families~\cite{parisi2019continual,delange2021survey,zhou2023cil}; Wang et
al.~\cite{wang2024comprehensive} find that rehearsal consistently
outperforms regularisation-only approaches as task dissimilarity increases.
\emph{Regularisation} methods such as Elastic Weight Consolidation
(EWC)~\cite{Kirkpatrick_2017} penalise changes to parameters important for
previous tasks; this limits weight change but not the decay of an
unreinforced feature, whose gradient can vanish without any weight change.
\emph{Architecture} methods (Progressive Neural
Networks~\cite{rusu2016progressive}, PackNet~\cite{mallya2018packnet})
allocate or mask per-task parameters to avoid forgetting by construction,
but scale poorly to long task sequences. \emph{Gradient-projection}
methods, Orthogonal Gradient Descent~\cite{farajtabar2019ogd} and Gradient
Projection Memory~\cite{saha2021gpm}, project new-task updates orthogonal
to previous-task directions, eliminating gradient \emph{interference}. That
assumption (forgetting as overwriting) only partly holds under the
scale-conditioned gradient imbalance examined in Section~\ref{sec:rq2},
where large-target capability degrades because the new distribution
supplies no positive gradient for it, not because it is overwritten.
\emph{Rehearsal} methods interleave stored exemplars with new data:
iCaRL~\cite{rebuffi2017icarl} selects exemplars by greedy herding to the
class mean (the strategy adapted for Scale-Stratified Herding here),
Chaudhry et al.~\cite{chaudhry2019tiny} show even tiny memories help,
GEM~\cite{lopezpaz2022gradientepisodicmemorycontinual} constrains updates
not to increase loss on stored exemplars, and Dark Experience
Replay~\cite{buzzega2020dark} adds logit distillation. Deng et
al.~\cite{deng_unlocking_2025} show rehearsal bounds forgetting when the
replayed distribution covers the original in feature space, motivating the
scale-stratified design of the SSH buffer.
Distillation methods use a frozen copy of the previous model as a
teacher~\cite{hinton2015distilling}; Learning without Forgetting
(LwF)~\cite{li2017lwf} penalises divergence from the teacher's output
distribution on new-task inputs without storing any previous data.
Knowledge distillation has also been investigated in multi-object tracking
outside the continual-learning setting. Faber et al. distilled DINOv2
representations into the FairMOT backbone using a teacher--student architecture
\cite{faber_leveraging_2024}. Their results showed scenario-dependent
improvements but did not consistently outperform the original FairMOT model.
This provides additional evidence that knowledge distillation can transfer
representational knowledge between vision models, while differing from the
present work, where distillation is used specifically to preserve detector
outputs across sequential domain shifts.

\subsection{Incremental Object Detection}

Applying continual learning to object detectors is non-trivial because
the detection loss couples classification, localisation, and objectness
objectives. Shmelkov et al.~\cite{shmelkov2017iod} established the
foundational approach: a distillation loss on the old-class output
logits of a Faster R-CNN prevents catastrophic forgetting when new classes
are added, while the detection loss on new-class ground truth drives
plasticity. Wang et al.~\cite{wang2022erd} refined this with Elastic
Response Distillation (ERD), which selectively weights the distillation
signal by the quality of each detection response, achieving
state-of-the-art incremental detection on MS COCO. Joseph et
al.~\cite{joseph2020iod} approached incremental object detection with a
meta-learning strategy that reshapes model gradients to minimise
forgetting while maximising knowledge transfer across incrementally added
classes. Both Shmelkov et al., ERD, and Joseph et al. study
class-incremental detection: new classes arrive sequentially on the same
image distribution. Song et al.~\cite{song2024diod} study
domain-incremental detection, the setting where the class set is fixed but
the image distribution changes across tasks, which is closer to the
protocol used in this research. Their non-exemplar method learns a compact
domain bias module per task, but does not quantify per-stratum forgetting
or examine gradient starvation as a mechanism. No prior work in incremental
object detection has (a) applied a CL protocol to a chain of real thermal
anti-UAV datasets or (b) identified scale-distribution shift between tasks
as a primary driver of forgetting.

\subsection{Gradient Starvation and Scale Distribution Shift}

Pezeshki et al.~\cite{pezeshki2021gradient} characterised gradient
starvation as a learning proclivity in which a dominant feature receives
the majority of the gradient signal, suppressing updates to weaker but
relevant features. This phenomenon has been studied in the context of
spurious correlation and shortcut learning in single-task settings, but
has not previously been linked to catastrophic forgetting in object
detection or anti-UAV surveillance. The connection this research draws is
that when a new task's training distribution lacks a feature class that
was dominant in previous tasks (specifically, large-target bounding boxes
in Anti-UAV-RGBT), the positive (target-matched) gradient signal for that
feature class largely vanishes during new-task training. Singh and Davis~\cite{singh2018snip}
demonstrated in the single-task setting that CNNs are not robust to scale
changes between training and test time: a model trained on large objects
underperforms systematically on small ones and vice versa. In the
continual learning setting studied here, this scale sensitivity manifests
across tasks rather than within a single evaluation: large-target features
learned on Anti-UAV-RGBT receive no reinforcing positive gradient signal
during CST Anti-UAV fine-tuning and decay as a result. Because the learnable weights are not strongly displaced in aggregate, the
resulting forgetting is dissociated from global weight-change magnitude, as captured by the inter-stage cosine similarity analysis in
Section~\ref{sec:rq2}. This distinguishes gradient starvation from the
gradient interference studied by methods such as OGD~\cite{farajtabar2019ogd}
and GPM~\cite{saha2021gpm}: those methods project conflicting gradients,
but cannot project gradients that do not exist. This scale-conditioned
imbalance is therefore a failure mode that existing regularisation,
gradient projection, and distillation methods are not designed to address,
since none restores a positive gradient signal that the new distribution does not
provide.

\section{Methodology}
\label{sec:methodology}

\begin{figure*}[t]
  \centering
  \includegraphics[width=0.9\textwidth]{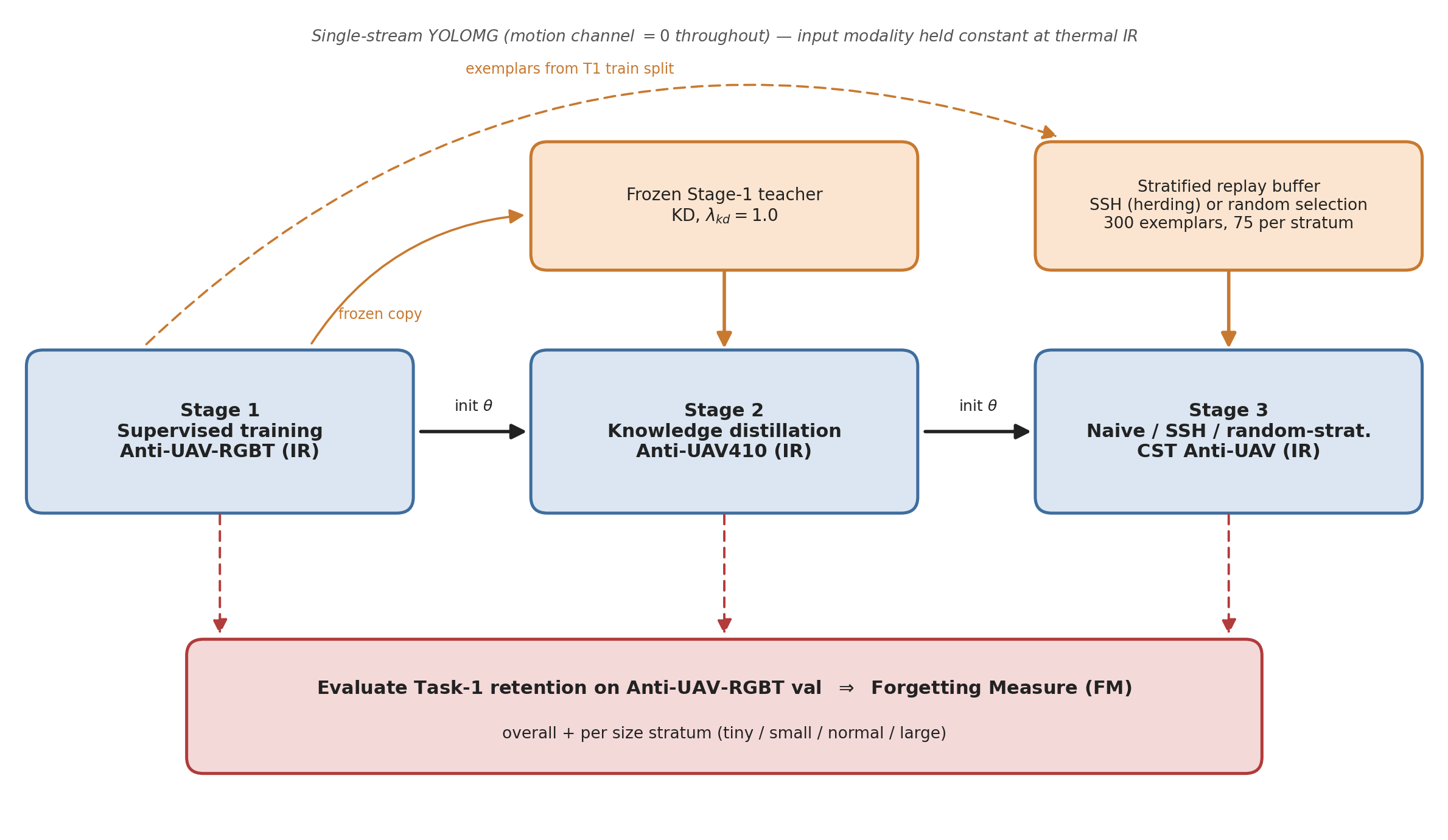}
  \caption{Overview of the three-stage continual-learning curriculum. A single
    YOLOMG detector, run as a single thermal-infrared stream with the motion
    channel zeroed, is trained sequentially: Stage~1 supervised on
    Anti-UAV-RGBT, Stage~2 with knowledge distillation from the frozen Stage~1
    teacher on Anti-UAV410, and Stage~3 (naive fine-tuning, Scale-Stratified
    Herding replay, or the random-stratified control) on CST Anti-UAV. The
    replay buffer is drawn from the Anti-UAV-RGBT training split (300
    exemplars, 75 per size stratum). After each stage, Task-1 retention is
    evaluated on the Anti-UAV-RGBT validation split to compute the Forgetting
    Measure, overall and per size stratum.}
  \label{fig:pipeline}
\end{figure*}

This research follows a sequential empirical evaluation design. A single
detector is trained across three anti-UAV datasets in order, and the
change in detection performance on earlier data is measured after each
subsequent training stage. The design is motivated by the continual
learning literature~\cite{wang2024comprehensive}, which formalises this
as a stability-plasticity problem: the model must retain prior knowledge
(stability) while acquiring new capabilities (plasticity). The protocol
corresponds to the domain-incremental learning scenario defined by Van de
Ven et al.~\cite{vandeven2022three}: the target class (UAV) is fixed
throughout all three stages, while the input distribution changes with
each new dataset. Unlike prior anti-UAV studies that evaluate each dataset
in isolation, this research chains Anti-UAV-RGBT, Anti-UAV410, and CST
Anti-UAV into a shared training history so that forgetting can be measured
directly. Figure~\ref{fig:pipeline} summarises the three-stage curriculum.

\subsection{Backbone: YOLOMG}

The detector is YOLOMG~\cite{guo_yolomg_2025}, a YOLOv5-based
architecture~\cite{jocher2020yolov5} extended with a second input
channel. The first channel is the infrared appearance frame. The second
channel is a pixel-level motion difference map (called mask32) computed
within a five-frame window, using the three frames $t{-}2$, $t$, $t{+}2$
(Appendix~\ref{sec:apx:preprocessing}), after global motion compensation (GMC).
GMC removes camera ego-motion using the Enhanced Correlation Coefficient
algorithm~\cite{evangelidis2008parametric}, leaving only object-relative
motion as the residual signal. The two channels are fused by
concatenation in the early convolutional layers. YOLOMG outputs
predictions at three detection heads (P3, P4, P5) for small, medium, and
large objects. Its anchors are small and dataset-specific (re-clustered for
tiny UAVs; largest $26\times14$\,px), not the YOLOv5 defaults. The full
model has 318 layers and 335 named parameter tensors.

YOLOMG was selected because the initial scope included motion-based
detection, with the mask32 channel intended to complement the infrared
stream~\cite{zhao_vision-based_2022,guo_yolomg_2025,zhu_evidential_2023}.
As the scope narrowed to a continual learning protocol, the motion channel
was ruled out: Anti-UAV-RGBT~\cite{jiang_anti-uav_2023} and
Anti-UAV410~\cite{bohuang_410} are appearance-based tracking datasets whose
inter-frame UAV displacement is too small for the GMC-based mask32 to yield
an informative signal. Rather than switch architecture mid-project, adding
a confound to the analysis, the motion channel was held at a zero tensor
throughout, making the model functionally a single-stream YOLOv5 appearance
detector while keeping the architecture consistent across the curriculum.

\subsection{Stage 1: Supervised Training on Anti-UAV-RGBT}

Stage 1 trains YOLOMG from ImageNet-pretrained weights on the infrared
stream of Anti-UAV-RGBT~\cite{jiang_anti-uav_2023} using fully
supervised detection. Ground-truth bounding box annotations are available
for the IR stream directly, so no domain adaptation is needed at this
stage. The model is trained for 100 epochs using SGD with momentum 0.937
and weight decay $5 \times 10^{-4}$, with a cosine learning rate
schedule decaying from an initial rate of $10^{-2}$. The best checkpoint
by validation mAP@0.5 is retained and used to initialise Stage 2. Stage
1 performance sets the T1 ceiling against which all subsequent Forgetting
Measure values are computed.

\subsection{Stage 2: Knowledge Distillation on Anti-UAV410}

Stage 2 fine-tunes the Stage 1 checkpoint on Anti-UAV410~\cite{bohuang_410},
a larger thermal infrared dataset that shares the same
large-to-normal scale regime as Stage 1 but covers a wider range of
acquisition conditions. The risk at this stage is that fine-tuning on
Anti-UAV410 destroys the representations built for Anti-UAV-RGBT. To
prevent this, Stage 2 uses knowledge distillation~\cite{hinton2015distilling,li2017lwf}:
the frozen Stage 1 checkpoint acts as a teacher, and the fine-tuning
student is penalised for diverging from the teacher's output
distribution.

The total training loss is:

\begin{equation}
L_{\text{total}} = L_{\text{det}} + \lambda_{\text{kd}} \cdot L_{\text{kd}}
\label{eq:kd_loss}
\end{equation}

where $L_{\text{det}}$ is the standard YOLOMG detection loss (box
regression, objectness, and classification) and $L_{\text{kd}}$ is the
mean squared error between the student and teacher prediction grids at the
three detection heads. The weight is fixed at $\lambda_{\text{kd}} = 1.0$,
weighting detection and distillation equally. This follows convention: both
Li and Hoiem~\cite{li2017lwf} and Shmelkov et al.~\cite{shmelkov2017iod}
use unit weighting as their default. Equal weighting is a neutral baseline,
biasing the optimisation toward neither plasticity (lower $\lambda$) nor
stability (higher $\lambda$) without task-specific tuning data. Stage 2 is run three times
with different random seeds (42, 123, 999) to obtain confidence intervals
on the Forgetting Measure and the peak T2 mAP.

\subsection{Stage 3: Naive Fine-tuning on CST Anti-UAV}

Stage 3 fine-tunes the best Stage 2 checkpoint on CST
Anti-UAV~\cite{xie2025cstantiuavthermalinfrared} without any memory
protection mechanism. This is the naive baseline against which future
mitigation strategies will be compared. The purpose of Stage 3 is
diagnostic. It quantifies how much forgetting the scale shift from
Anti-UAV410 (49\% normal/large) to CST (99.8\% tiny/small) causes, and
examines the parameter-space changes that accompany it. Training runs for
the number of epochs needed to reach peak T3 mAP, after which T1 is evaluated.

\subsection{Replay Design: Scale-Stratified Herding}

The gradient-starvation interpretation developed in the Stage 3 analysis
motivates the design of Scale-Stratified Herding (SSH). Standard iCaRL
herding~\cite{rebuffi2017icarl} selects exemplars whose penultimate-layer
embeddings are closest to the class mean, but treats all samples
uniformly. In the anti-UAV setting, the T1 training split is imbalanced
across size strata (Table~\ref{tab:buffer_counts}). A flat herding buffer
would replicate this imbalance rather than guarantee coverage of every
stratum, so the strata that are scarce in T1 would receive little gradient
protection during Stage 3.

SSH partitions the buffer capacity equally across four size strata (tiny,
small, normal, large) defined by bounding box area in pixels. Exemplars
are drawn from the Anti-UAV-RGBT training split (148{,}368 UAV-present
frames), kept disjoint from the validation set used to compute the FM. Within each stratum, greedy
herding selects the 75 samples closest to the stratum mean embedding, for
a total buffer of 300 exemplars. This small fixed memory follows the
low-budget rehearsal convention of iCaRL~\cite{rebuffi2017icarl} and the
finding that even tiny episodic memories curb
forgetting~\cite{chaudhry2019tiny}; the equal 75-per-stratum split is the
neutral choice that removes the source-distribution imbalance. As the
qualitative conclusion (stratified replay reduces forgetting relative to
naive fine-tuning) does not depend on the exact budget, a systematic
buffer-size sweep is left to future work. During Stage 3, buffer samples are
replayed at one exemplar per four CST samples per mini-batch. The replayed
distribution therefore spans the full T1 scale regime, intended to
counteract the large-target gradient suppression examined in the S2$\to$S3
analysis. To
separate the effect of the herding selection from that of the stratification
itself, SSH is compared both against naive fine-tuning and against a
random-stratified control: the same buffer size and per-stratum quotas, but
random selection within each stratum in place of herding
(Section~\ref{sec:rq3}).

\subsection{Datasets}

Four datasets are used across the three stages. Anti-UAV-RGBT provides
318 paired RGB and IR video sequences with per-frame bounding box
annotations on the IR stream. Across all splits it contains 293,209
UAV-present frames (frames in which a UAV is present, the \texttt{exist}=1
flag); its training and validation splits (208,988 frames) have a scale
distribution dominated by normal targets: 0.5\% tiny, 28.0\% small,
67.8\% normal, and 3.7\% large. Anti-UAV410 contains 428,703 UAV-present
frames across all splits, with a train and validation scale distribution of
24.8\% tiny, 25.9\% small, 46.6\% normal, and 2.7\% large. CST Anti-UAV
contains 208,221 UAV-present frames across all splits, with a near-inverted
train and validation distribution: 87.9\% tiny, 11.9\% small, 0.3\% normal,
and 0\% large. ARD100 provides 199,291 UAV-present ego-motion frames
(202,411 total) used exclusively to pre-compute the motion difference masks
for the YOLOMG temporal branch and is not part of the continual learning
curriculum.

All datasets are accessed from their native storage formats during
training to stay within the Snellius inode quota. Video-based datasets
(Anti-UAV-RGBT, ARD100) are decoded on demand using cv2.VideoCapture;
image-based datasets (Anti-UAV410, CST) are read from pre-extracted JPEG
frames. Bounding box coordinates are normalised to $[0,1]$ relative to a
$640 \times 512$ resolution.

\subsection{Evaluation Metrics}

The Forgetting Measure (FM) is the primary metric for quantifying
retention of earlier-task performance~\cite{chaudhry2019tiny}:

\begin{equation}
\text{FM} = \text{mAP}_{T1}^{\text{after } T_k} - \text{mAP}_{T1}^{\text{after } T1}
\label{eq:fm}
\end{equation}

where both evaluations use the Anti-UAV-RGBT validation split. A negative FM
indicates forgetting. The T1 ceiling is the Stage 1 best-checkpoint mAP
(0.6725). Per-stratum mAP decomposes detection across the four size bins by
bounding-box area (thresholds 256, 1024, and 4096\,px$^2$, i.e.\ $16^2$,
$32^2$, and $64^2$). The scale-distribution figures, the per-stratum mAP,
and the replay buffer all use this single area convention. Parameter
drift between consecutive checkpoints is the mean cosine similarity across
all 335 named tensors. Values near 1.0 indicate near-static weights,
pointing to gradient starvation rather than bulk weight collapse.

For tracking evaluation on CST, Success Rate at Intersection over Union
(IoU) 0.5 (SR@0.5), Precision Rate at 20 pixels (PR@20), and Identity
Switch count (IDSW)
are reported using the one-pass evaluation protocol~\cite{bohuang_410}.

\subsection{Compute Infrastructure}
\label{sec:compute}

All experiments are run on Snellius (SURF national supercomputer), using
the A100 and H100 GPU partitions (per-stage allocation in
Table~\ref{tab:apx:config}). The full three-stage protocol---including the
three Stage 2 seed replicates and the three Stage 3 variants---required on
the order of one hundred hours of wall-clock training on 4-GPU nodes. This
compute scale frames the scope of the study: no on-board or edge
deployment is assumed, and whether resource-constrained hardware could
support the continual (re)training studied here was not investigated;
on-device continual learning is an open research problem in its own
right~\cite{pellegrini2021edge,park2025replay}. The Python environment uses Python 3.9, PyTorch 2.7.1,
and CUDA 11.8. Random seeds are fixed at 42, 123, and 999 across all
libraries to ensure reproducibility of multi-seed Stage 2 runs.

\section{Results}
\label{sec:results}

\subsection{Scale-Distribution Shift Across Domains}
\label{sec:scale_dist}

\begin{figure}[h]
  \centering
  \includegraphics[width=\linewidth]{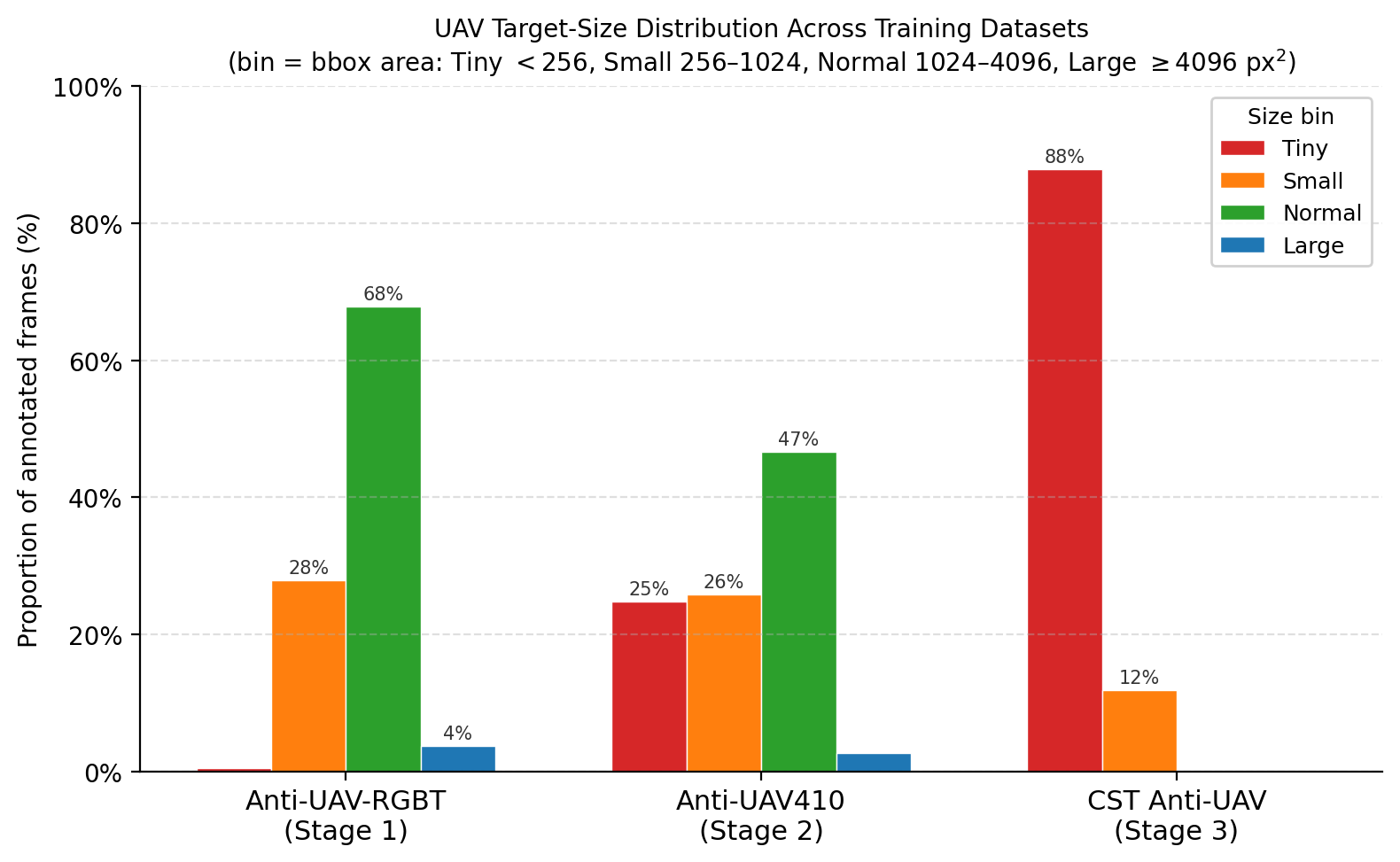}
  \caption{Scale distribution across the three training datasets (train +
    val splits, bbox area: tiny $<$256\,px$^2$, small 256--1024\,px$^2$,
    normal 1024--4096\,px$^2$, large $\geq$4096\,px$^2$). Anti-UAV-RGBT is
    dominated by normal targets (67.8\%; 95.8\% normal or small). CST
    Anti-UAV shows a near-complete inversion: 87.9\% tiny; 99.8\% tiny
    or small. All three datasets are thermal infrared, so the modality is
    held constant; the shift studied here is this scale/size distribution,
    not an arbitrary domain shift.}
  \label{fig:sizes}
\end{figure}

Before reporting per-stage results it is necessary to characterise how
severely the three training domains differ in target size, since this
difference is the primary driver of the forgetting observed in Stage 3.
Figure~\ref{fig:sizes} shows the percentage of UAV-present frames in each
of four UAV-specific size categories across all datasets. Size boundaries
are by bounding-box area: tiny $<$256\,px$^2$, small 256--1024\,px$^2$,
normal 1024--4096\,px$^2$, large $\geq$4096\,px$^2$ (i.e.\ $16^2$ to
$64^2$). Standard COCO thresholds
(32/96/192\,px) do not suit UAV detection. A drone at 300\,m typically
subtends fewer than 32 pixels in thermal imagery. Under COCO conventions,
then, most anti-UAV targets would count as ``small'' or ``tiny''
regardless of dataset.

In Anti-UAV-RGBT (T1), 95.8\% of UAV-present frames are normal or small
(67.8\% normal, 28.0\% small) and only 3.7\% are large; in CST Anti-UAV
(T3), the normal and large strata are nearly absent (0.3\% and 0\%) and
tiny targets dominate at 87.9\%. Anti-UAV410 (T2) is intermediate (46.6\%
normal, 24.8\% tiny), providing partial distribution overlap with T1. This
inversion echoes Singh and
Davis~\cite{singh2018snip}: detectors trained on one scale regime perform
poorly at a different scale. Their study was single-task; here the effect
appears across sequential stages. The inversion motivates the gradient
analysis that follows and the Scale-Stratified Herding design.

\subsection{Catastrophic Forgetting Under Naive Sequential Training}
\label{sec:h1}

Naive sequential training is expected to induce measurable catastrophic
forgetting of T1 performance. The Forgetting Measure is:

\begin{equation}
  \mathrm{FM}_k = \mathrm{mAP}_{T1}^{\,\text{after }T_k}
    - \mathrm{mAP}_{T1}^{\,\text{after }T1}
  \label{eq:fmk}
\end{equation}

where $\mathrm{mAP}_{T1}$ is evaluated on the Anti-UAV-RGBT validation split
and a value of zero indicates perfect retention.

Training the Stage 2 checkpoint on CST Anti-UAV for 19 epochs without any
replay buffer drives T1 mAP@0.5 from 0.640 (the post-Stage-2 level) down
to 0.068 at the best T3 checkpoint (epoch 3), and to 0.017 at the final
epoch. Against the Stage 1 ceiling (0.6725) this is
$\mathrm{FM} = -0.605$. Of this, $-0.572$ is incurred in Stage 3 itself
($\mathrm{FM}_{\text{stage3}}$, relative to the Stage 2 baseline); the
remaining $-0.033$ was already lost during Stage 2 distillation. The
absolute post-Stage-3 forgetting ($-0.605$) is thus roughly 18 times the
Stage 2 forgetting ($\mathrm{FM} = -0.033 \pm 0.004$ across three seeds),
and the Stage 3 increment alone ($-0.572$) about 17 times. Under these stage-specific training protocols, the Stage~3 transition is
associated with far more forgetting than the Stage~2 transition. This ratio
must not be interpreted as the isolated causal effect of scale shift because
KD is used in Stage~2 but not in the naive Stage~3 condition.

Table~\ref{tab:stratum_forgetting} shows the per-stratum breakdown.
Large-target T1 mAP collapses to 0.000 by the end of training. The
forgetting is concentrated precisely in the large-target stratum that CST
Anti-UAV lacks entirely, while the tiny stratum is barely affected.

\begin{table}[h]
  \centering
  \caption{Per-stratum T1 mAP@0.5 on Anti-UAV-RGBT across training stages.
Stage 3 values are from the best naive checkpoint (epoch 3). After S$k$ is
    T1 mAP after Stage $k$; $\Delta$ is the net Stage 1 to Stage 3 change.
    Knowledge distillation retains every stratum through Stage 2---the large
    stratum is even slightly strengthened (0.461 to 0.494)---and the collapse
    occurs only at Stage 3. After S2 is the mean over three Stage 2 seeds
    (per-stratum std $\leq 0.006$); After S1 (the Stage 1 ceiling) and After
    S3 (the naive run) are single checkpoints. Strata are defined by
    bounding-box area (thresholds 256/1024/4096\,px$^2$).}
  \label{tab:stratum_forgetting}
  \small
  \begin{tabular}{@{}lrrrr@{}}
    \toprule
    Stratum & After S1 & After S2 & After S3 & $\Delta$ \\
    \midrule
    Tiny   & 0.009 & 0.018 & 0.001 & $-$0.008 \\
    Small  & 0.579 & 0.559 & 0.060 & $-$0.519 \\
    Normal & 0.719 & 0.684 & 0.079 & $-$0.640 \\
    Large  & 0.461 & 0.494 & 0.000 & $-$0.461 \\
    \midrule
    Overall             & 0.673 & 0.640 & 0.068 & $-$0.605 \\
    \bottomrule
  \end{tabular}
\end{table}

\begin{figure*}[t]
  \centering
  \includegraphics[width=\textwidth]{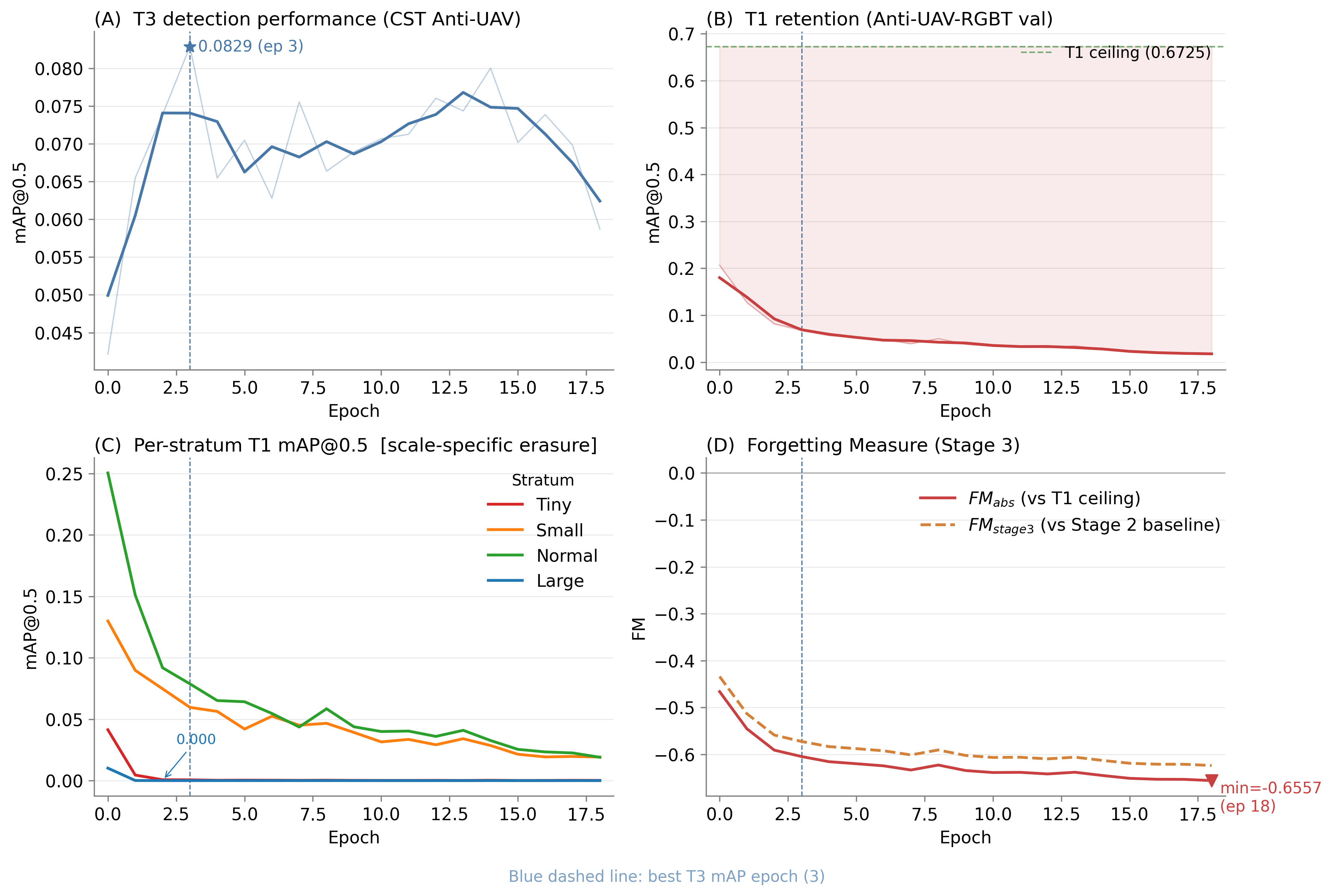}
  \caption{Stage 3 naive fine-tuning on CST Anti-UAV (19 epochs). (A) T3
    detection mAP@0.5 on CST val; star marks the best checkpoint at epoch
    3 (0.083). (B) T1 retention on Anti-UAV-RGBT val; shaded region shows
    the gap to the Stage 1 ceiling (0.6725). (C) Per-stratum T1 mAP@0.5:
    the large stratum (green) collapses to 0.000 at epoch 1, indicating
    scale-specific erasure. (D) Forgetting Measure: FM$_{\text{abs}}$
    (vs Stage 1 ceiling, dashed) and FM$_{\text{stage3}}$ (vs Stage 2
    baseline, solid); worst FM$_{\text{abs}} = -0.656$ at epoch 18.}
  \label{fig:stage3}
\end{figure*}

The per-epoch dynamics of the naive condition are shown in
Figure~\ref{fig:stage3} and detailed in
Appendix~\ref{sec:apx:dynamics} (Table~\ref{tab:s3_naive_epochs}). The
large stratum falls to near-zero within the first training epoch and to
0.000 by epoch~1, never recovering. The normal stratum, the model's primary
detection strength after
Stage 2 (mAP 0.681 on Anti-UAV410, Table~\ref{tab:stage2_scale}), collapses
from 0.251 at epoch~0 to 0.044 by epoch~7 and to 0.019 by the final epoch.
T3 mAP peaks at epoch~3 (0.083) and then plateaus while T1 mAP continues
declining; forgetting does not stop when the new task stops improving, a
hallmark of catastrophic forgetting~\cite{mccloskey1989catastrophic}.

The forgetting is therefore measurable, large, and concentrated by
stratum.

\subsection{Knowledge Distillation During Stage 2 Training}
\label{sec:rq1}

Stage 2 training combines a standard detection loss with a knowledge
distillation (KD) term to limit forgetting of Task 1 capability. The combined loss is:

\begin{equation}
  L_{\text{total}} = L_{\text{det}} + \lambda_{\text{kd}} \cdot L_{\text{kd}},
  \qquad \lambda_{\text{kd}} = 1.0
  \label{eq:stage2_loss}
\end{equation}

where $L_{\text{det}}$ is the standard detection loss on Anti-UAV410
labels and $L_{\text{kd}}$ is the mean squared error between the student
and the frozen Stage 1 teacher at prediction heads P3, P4, P5.

Table~\ref{tab:stage2_agg} reports Stage 2 aggregate results across three
random seeds (42, 123, 999). KD limits forgetting to 2.8--3.7 percentage
points of T1 mAP across the three seeds (FM $-0.033 \pm 0.004$).

\begin{table}[h]
  \centering
  \caption{Stage 2 aggregate results: KD fine-tuning on Anti-UAV410
    ($\lambda_{\text{kd}} = 1.0$, mean $\pm$ std across three seeds).}
  \label{tab:stage2_agg}
  \begin{tabular}{@{}lr@{}}
    \toprule
    Metric & Value \\
    \midrule
    T2 mAP@0.5 on Anti-UAV410 (best epoch) & 0.428 $\pm$ 0.002 \\
    T1 mAP@0.5 retained (Anti-UAV-RGBT)    & 0.640 \\
    Forgetting Measure FM                   & $-$0.033 $\pm$ 0.004 \\
    T1 retention                            & 95\% \\
    Best T2 epoch (seed 42)                 & 16 / 32 \\
    \bottomrule
  \end{tabular}
\end{table}

A per-stratum breakdown indicates this retention is genuine, not an
aggregate artefact (Table~\ref{tab:stratum_forgetting}): every T1 stratum
is preserved through Stage 2, and the large stratum, the one CST later
erases, is even slightly strengthened (0.461 to 0.494). The 95\% figure
therefore reflects true preservation of the T1 representation, not mere
overlap between Anti-UAV410 and T1.

\begin{figure*}[t]
  \centering
  \includegraphics[width=\textwidth]{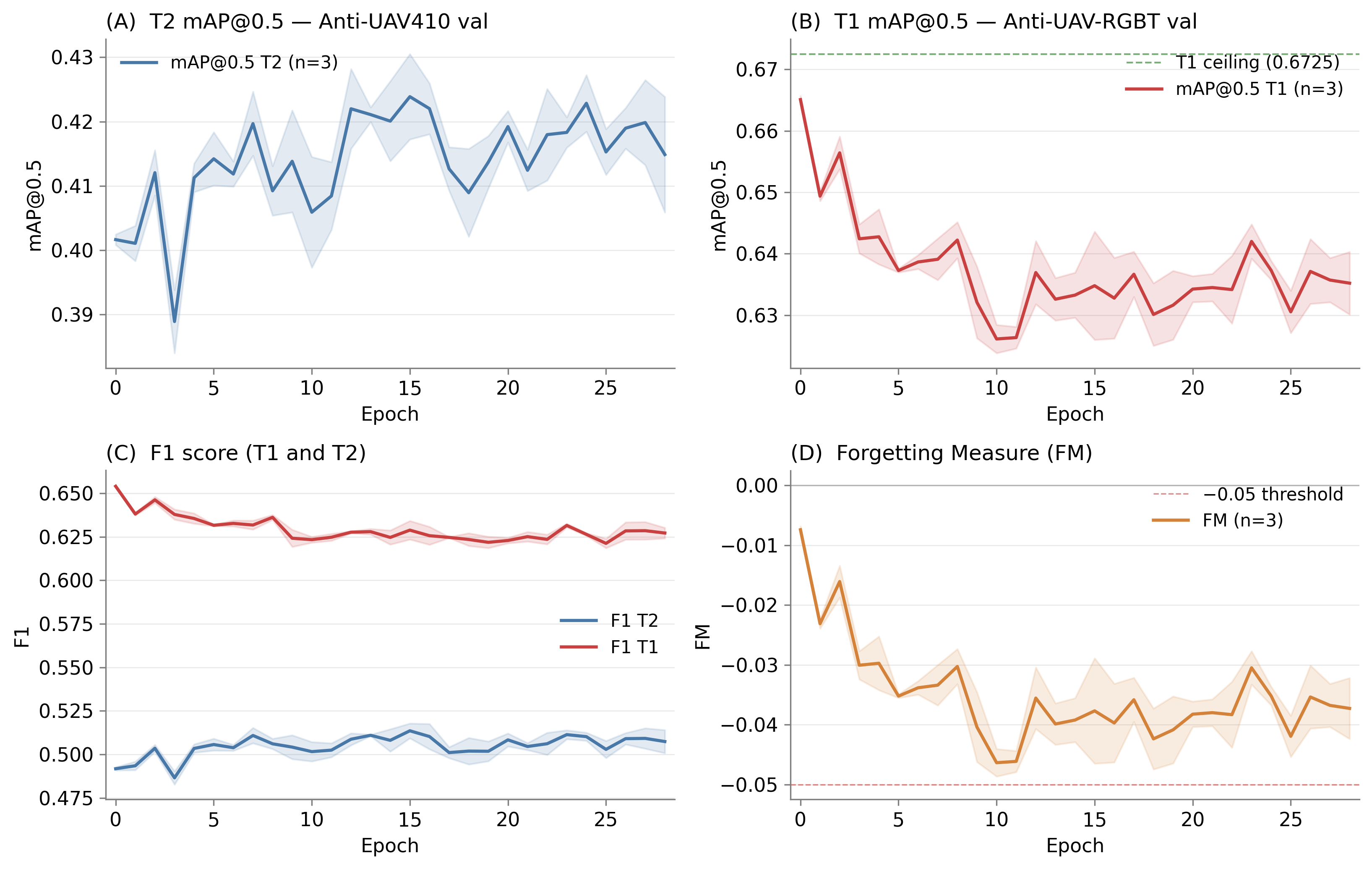}
  \caption{Stage 2 training dynamics across three seeds (n=3, mean with
    shaded $\pm$1\,std). (A) T2 mAP@0.5 on Anti-UAV410 val. (B) T1
    mAP@0.5 on Anti-UAV-RGBT val; green dashed line is the Stage 1
    ceiling (0.6725). (C) F1 score for T1 and T2 tasks. (D) Forgetting
    Measure per epoch; dashed horizontal marks the $-$0.05 boundary.}
  \label{fig:stage2}
\end{figure*}

Figure~\ref{fig:stage2} shows the per-epoch Stage 2 trajectories across the
three seeds. Epoch-level dynamics for seed 42 (Appendix~\ref{sec:apx:dynamics},
Table~\ref{tab:stage2_epochs}) show two patterns. First, the kd/det ratio
rises from 1.36 at epoch 0 to 2.62 at epoch 31: $L_{\text{det}}$ declines
faster than $L_{\text{kd}}$ as the student adapts to Anti-UAV410, after
which distillation provides a growing share of the gradient signal. Second,
the FM deepens from $-$0.007 at epoch 0 to a trough of $-$0.046 at epoch 10,
then partially recovers to $-$0.037 at the peak T2 epoch (16), illustrating
the stability-plasticity trade-off directly: the checkpoint with the
highest T2 mAP is not the one with the smallest forgetting.

The mean cosine similarity between Stage 1 and Stage 2 best checkpoints
across all 335 named parameter tensors is 0.911 (per-seed values: 0.917,
0.921, 0.896), indicating high overall weight alignment. Restricting the
measure to the 204 learnable (gradient-updated) tensors---excluding
BatchNorm running-statistic buffers, which adapt by forward-pass averaging
rather than backpropagation---gives a comparable 0.935 (seed~42), so the
alignment is a property of the trained weights rather than an artefact of
the buffer count. The spatial attention module shows the largest relative
drift, indicating the model adapted its low-level attention patterns for
Anti-UAV410 while keeping backbone features close to the teacher.

Table~\ref{tab:stage2_scale} reports the scale-stratified mAP@0.5 on the
Anti-UAV410 validation set (mean across three seeds). The Stage 2 model
inherits the Stage 1 detector's strength at the normal stratum
and retains moderate small-target performance, but is essentially unable
to detect tiny targets ($<$256\,px$^2$). This profile anticipates the
scale-conditioned forgetting pattern of Stage 3: when fine-tuning on CST
Anti-UAV (99.8\% tiny/small, 0\% large), the normal and large strata, almost
absent from CST, receive little positive gradient signal from the new
training distribution, a pattern probed directly in
Section~\ref{sec:gradprobe}.

\begin{table}[h]
  \centering
  \caption{Stage 2 scale-stratified mAP@0.5 on Anti-UAV410 validation set
    (mean across seeds 42, 123, 999). Normal-stratum performance (0.681) is
    the model's primary strength; tiny-stratum performance (0.001) is
    essentially zero throughout all stages.}
  \label{tab:stage2_scale}
  \begin{tabular}{@{}lcc@{}}
    \toprule
    Stratum & Area range & mAP@0.5 (mean) \\
    \midrule
    Tiny   & $<$256\,px$^2$         & 0.001 \\
    Small  & 256--1024\,px$^2$      & 0.278 \\
    Normal & 1024--4096\,px$^2$     & \textbf{0.681} \\
    Large  & $\geq$4096\,px$^2$     & 0.276 \\
    \bottomrule
  \end{tabular}
\end{table}

Together, these results show that Stage~2 training with KD at
$\lambda_{\text{kd}} = 1.0$ retains 95\% of the Stage~1 T1 mAP
($\text{FM} = -0.033 \pm 0.004$). The rising kd/det ratio and high inter-stage
cosine similarity describe the optimisation behaviour, but a no-KD Stage~2
control would be required to quantify the causal contribution of KD.

\subsection{Scale Shift and the Forgetting Pattern in Stage 3}
\label{sec:rq2}

CST Anti-UAV's training split contains 99.8\% tiny/small targets and 0\%
large targets. When the Stage 2 checkpoint is fine-tuned on CST without
replay, large-target T1 mAP falls to near-zero within the first training
epoch and to 0.000 by epoch~1, while the tiny stratum barely changes
(0.001 to 0.000). The collapse is not distributed uniformly: strata absent from T3 training
data receive no positive target-matched gradient signal and collapse first.

To probe what underlies this pattern, inter-stage parameter drift is
compared at both transitions (Table~\ref{tab:cosine_sim}).

\begin{table}[h]
  \centering
  \caption{Mean cosine similarity between successive stage checkpoints,
    computed over all 335 named tensors and, separately, over the 204
    learnable (gradient-updated) tensors only, with BatchNorm
    running-statistic buffers excluded. Learnable-only values are for
    seed~42. Higher value means smaller weight change. FM is the
    Forgetting Measure relative to the Stage~1 ceiling.}
  \label{tab:cosine_sim}
  \small
  \begin{tabular}{@{}lccc@{}}
    \toprule
    Transition & Cosine (all) & Cosine (learn.) & FM \\
    \midrule
    Stage 1 to Stage 2 & 0.911 & 0.935 & $-$0.033 \\
    Stage 2 to Stage 3 & 0.967 & 0.987 & $-$0.605 \\
    \bottomrule
  \end{tabular}
\end{table}

\begin{figure}[h]
  \centering
  \includegraphics[width=\linewidth]{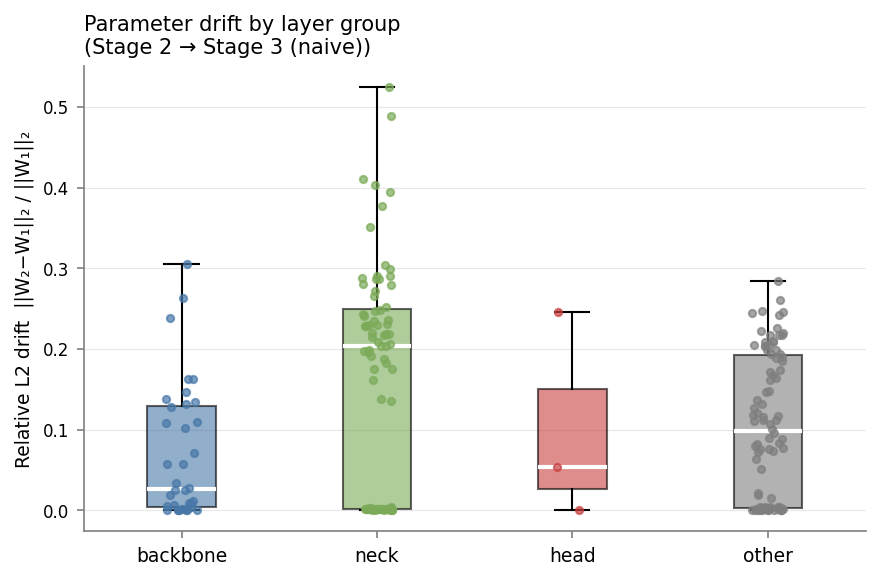}
  \caption{Relative L2 parameter drift (Stage 2 $\to$ Stage 3, naive) per
    layer group, restricted to the learnable (gradient-updated) tensors.
    The neck adapts most (median $\approx$0.20) and the backbone least
    (median $\approx$0.03); the three detection-head tensors show only
    modest drift (median $\approx$0.05). Excluding the BatchNorm
    running-statistic buffers removes the spurious head-layer drift seen
    when all parameters are pooled: the gradient-updated head weights
    barely move, consistent with starvation of the absent large-target
    stratum rather than active overwriting of the output layers.}
  \label{fig:drift_groups}
\end{figure}

The cosine similarity increases between transitions whether measured over
all parameters (0.911 to 0.967) or over the learnable, gradient-updated
tensors alone (0.935 to 0.987; Table~\ref{tab:cosine_sim}). By either
measure the trained weights moved \emph{less} in Stage 3 than in Stage 2,
relative to their starting point, yet FM is 18 times larger. This
decoupling of weight-change magnitude from forgetting magnitude is
consistent with gradient starvation~\cite{pezeshki2021gradient}:
large-target representations appear to receive insufficient gradient
signal during Stage 3 training, as CST contains no large-target examples.
The drift that does occur in Stage 3 sits not in the gradient-updated
weights but in the BatchNorm running statistics (relative L2 $\approx$0.24
for \texttt{running\_mean} and \texttt{running\_var}, versus $\approx$0.12
for the learnable weights); among the learnable tensors the drift
concentrates in the neck rather than the detection heads
(Figure~\ref{fig:drift_groups}). The network tracks CST's input distribution
through its normalisation buffers, while the trained weights stay close to
their Stage 2 values. The unreinforced large-target output regime may then degrade under the
combined influence of weight decay and the small-target box-regression terms
that dominate CST.
The result is a stratum-specific collapse concentrated in the large
stratum, the only stratum at 0.000 from epoch~1 onward, while the tiny
stratum, already near zero in T1, is essentially unaffected.

Taken together, these results indicate that scale-distribution shift
characterises the Stage 3 forgetting pattern at the stratum level. The
evidence is consistent with a scale-conditioned gradient mechanism:
inter-stage cosine similarity is higher at Stage 3 than at Stage 2 (0.987
vs 0.935 on the learnable tensors; 0.967 vs 0.911 on all parameters), yet
forgetting is 18 times greater. Section~\ref{sec:gradprobe} probes this
mechanism directly.

\subsection{Direct Gradient Probe at the Stage 2\,$\to$\,3 Boundary}
\label{sec:gradprobe}

The cosine and BatchNorm evidence above is indirect. To examine the
gradient signal directly, the Stage~2 checkpoint was probed \emph{before}
any Stage~3 optimisation: for 60 mini-batches each of CST (the Stage~3
distribution) and Anti-UAV-RGBT (T1), a forward and backward pass was run
\emph{without an optimiser step} (the network in training mode for loss
computation, with BatchNorm layers frozen in evaluation mode so their
running statistics were not perturbed), and the $L_2$ norm of the gradient
reaching each detection head was recorded, together with the
box-regression loss $L_{\text{box}}$. The probe was repeated for all three
Stage~2 seeds; the values are stable across seeds and are reported as
seed means in Table~\ref{tab:gradprobe}.

\begin{table}[h]
  \centering
  \caption{Direct gradient probe at the Stage~2\,$\to$\,3 boundary.
    Per-head gradient $L_2$ norm and box-regression loss $L_{\text{box}}$
    (mean over three Stage~2 seeds, 60 batches each, no optimiser step),
    under the CST distribution, the full T1 distribution, and T1 with only
    large ($\geq$4096\,px$^2$) ground truth retained. Heads are ordered by anchor
    size (P3: $\sim$3\,px; P5: up to $26\times14$\,px).}
  \label{tab:gradprobe}
  \small
  \begin{tabular}{@{}lrrr@{}}
    \toprule
    Quantity & CST (0\% large) & RGBT (all) & RGBT (large only) \\
    \midrule
    $\|g\|$ P3 (fine)   & 0.510 & 0.000 & 0.000 \\
    $\|g\|$ P4 (mid)    & 0.278 & 0.426 & 0.000 \\
    $\|g\|$ P5 (coarse) & 0.388 & 0.615 & 0.227 \\
    \midrule
    $L_{\text{box}}$    & 0.096 & 0.020 & 0.001 \\
    \bottomrule
  \end{tabular}
\end{table}

Three observations follow. First, gradient flow is strongly
scale-conditioned: a head receives gradient in proportion to the presence
of size-matched targets. The clearest case is the finest head ($\sim$3\,px
anchors), essentially silent on T1 ($\|g\| \approx 0.000$, as RGBT contains
almost no tiny ($<$256\,px$^2$) targets) but strongly driven on CST ($\|g\| \approx 0.51$,
tiny-dominated). This is consistent with the starvation principle: a
stratum absent from the training distribution receives no reinforcing
signal.

Second, the detector's anchors are small by design (largest
$26\times14$\,px). A $\geq$64\,px target is therefore weakly matched under
the anchor ratio test, producing a near-zero positive box signal \emph{even
in T1} ($L_{\text{box}} \approx 0.001$ on large-only batches versus
$0.020$ on full T1). Large-target detection is therefore a fragile
extrapolation rather than an anchor-matched capability, consistent with
its rapid collapse once the distribution shifts. This capability was
nonetheless real after Stage~1 (large-stratum mAP@0.5 = 0.461,
Table~\ref{tab:stratum_forgetting}): the probe explains why it is not
maintained under CST, not that it was never present.

Third, CST drives a box-regression loss roughly five times larger than T1
($L_{\text{box}} \approx 0.096$ vs $0.020$): Stage~3 applies strong
small-target localisation gradients to the shared heads while the large
stratum receives no positive reinforcement, since CST's small targets keep
even the coarsest head active ($\|g\| \approx 0.39$). The forgetting is
thus better described as a \emph{scale-conditioned gradient imbalance}
(absence of positive large-target signal combined with dominant
small-target gradients) than as starvation of a single dedicated head.

These measurements refine, rather than overturn, the cosine and BatchNorm
evidence of Section~\ref{sec:rq2}: the learnable weights move little
globally (Table~\ref{tab:cosine_sim}), while the large-target output
regime, never strongly anchored and now unreinforced, is not maintained.
The probe supports the scale-conditioned account as a \emph{candidate}
mechanism; isolating its causal weight from the concurrent batch-size and
learning-rate differences (Section~\ref{sec:discussion}) would require a
controlled Stage~3 rerun and is left to future work.

\subsection{Scale-Stratified Replay and Stage 3 Forgetting}
\label{sec:rq3}

Scale-Stratified Herding (SSH) addresses Stage 3 large-target forgetting by
re-injecting large-target gradient signal through stratified exemplar
replay, targeting the scale gap identified above.

The SSH buffer contains 300 exemplars partitioned equally across four
size strata (75 per stratum). Within each stratum, greedy iCaRL
herding~\cite{rebuffi2017icarl} selects the 75 samples whose
penultimate-layer embeddings are closest to the stratum mean, computed
using the Stage 2 best checkpoint as the feature extractor:

\begin{equation}
  x_k = \arg\min_{x \notin \mathcal{B}_{k-1}}
    \left\| \mu_s - \frac{1}{k}
    \left( \sum_{j<k} e(x_j) + e(x) \right) \right\|_2
\end{equation}

\noindent where $e(\cdot)$ is the global-average-pooled neck (P3) feature
of the Stage 2 encoder and $\mu_s$ is the mean embedding of stratum $s$,
matching the notation of Algorithm~\ref{alg:ssh}.

During Stage 3, each mini-batch mixes CST samples with replay exemplars
from the buffer, restoring the large-target gradient signal CST cannot
provide. This follows Deng et al.~\cite{deng_unlocking_2025}: rehearsal is
effective when the replayed distribution matches the original task in
feature space. Stratifying by scale covers the full T1 scale range, whereas
flat herding would replicate the training-split frequencies
(Table~\ref{tab:buffer_counts}), roughly two-thirds from the normal
stratum alone, and leave the scarce strata uncovered.

The buffer is built using the Stage~2 seed-42 checkpoint as the feature
extractor on the Anti-UAV-RGBT \emph{training} split (148{,}368 UAV-present
frames), kept disjoint from the validation set used to compute the FM so
that replay introduces no train--evaluation overlap.
Table~\ref{tab:buffer_counts} shows the per-stratum counts. The tiny
stratum (1{,}091 eligible frames) is sampled at 6.9\%, several times the
large-stratum rate and 30--70$\times$ the small- and normal-stratum rates,
deliberately over-representing tiny targets to protect that rare knowledge
during Stage~3.

\begin{table}[h]
  \centering
  \caption{SSH buffer composition (Anti-UAV-RGBT training split). Each
    stratum contributes exactly 75 exemplars regardless of size; the tiny
    stratum is over-represented (6.9\% sampling rate) to protect rare
    tiny-target knowledge. Strata are by bounding-box area
    (thresholds 256/1024/4096\,px$^2$).}
  \label{tab:buffer_counts}
  \begin{tabular}{@{}lrrr@{}}
    \toprule
    Stratum & Eligible frames & Selected & Sampling rate \\
    \midrule
    Tiny   &   1{,}091 & 75 & 6.9\% \\
    Small  &  43{,}560 & 75 & 0.2\% \\
    Normal &  98{,}929 & 75 & 0.1\% \\
    Large  &   4{,}788 & 75 & 1.6\% \\
    \midrule
    \textbf{Total}      & 148{,}368 & \textbf{300} & --- \\
    \bottomrule
  \end{tabular}
\end{table}

Trained from the Stage~2 seed-42 checkpoint, both replay conditions
substantially mitigate the Stage~3 collapse (Table~\ref{tab:ssh_results}).
At their best-T3 checkpoints (epoch~2), SSH reaches a Forgetting Measure of
$-0.311$ and random-stratified replay $-0.221$, against the naive $-0.605$:
each roughly halves the forgetting and holds large-target T1 mAP non-zero
(0.079 and 0.129 respectively, versus 0.000), at a small cost to T3
detection (0.064 vs the naive 0.083 mAP), the expected
stability--plasticity trade-off of replay.

\begin{table}[h]
  \centering
  \caption{Stage 3 forgetting: naive fine-tuning versus random-stratified
    replay and Scale-Stratified Herding (SSH), each at its best-T3
    checkpoint (single seed, seed 42). Both replay variants roughly halve
    the Forgetting Measure and keep the large stratum non-zero; herding does
    not improve on random selection.}
  \label{tab:ssh_results}
  \small
  \begin{tabular}{@{}lcccc@{}}
    \toprule
    Condition & T3 mAP & T1 mAP & FM & Large T1 \\
    \midrule
    Naive (ep.~3)          & 0.083 & 0.068 & $-$0.605 & 0.000 \\
    Random-strat.\ (ep.~2) & 0.064 & 0.451 & $-$0.221 & 0.129 \\
    SSH (ep.~2)            & 0.064 & 0.362 & $-$0.311 & 0.079 \\
    \bottomrule
  \end{tabular}
\end{table}

The random-stratified control isolates the source of the gain. Random
selection within each stratum matches---indeed slightly exceeds---greedy
herding, so the benefit appears to come from the \emph{scale
stratification} (balanced coverage of all size strata), not from the
iCaRL-style herding selection. In this setting the herding component
therefore seems unnecessary. Both runs are single-seed (seed 42) and short
(Section~\ref{sec:discussion}); multi-seed, longer-schedule replication is
left to future work.

\subsection{Tracking Baseline on CST Anti-UAV}
\label{sec:tracking}

To establish a detection-as-tracking baseline, the Stage 3 naive
checkpoint was evaluated on all 40 CST Anti-UAV validation sequences
(39,055 ground-truth frames) under the one-pass protocol; the full
precision and success plots are given in Appendix~\ref{sec:apx:tracking}.
Tracking quality is low, as expected from the detector's collapse on CST:
SR@0.5 = 0.012, AUC-SR = 0.039, PR@20 = 0.219, and 135 identity switches.
The gap between PR@20 and SR@0.5 suggests the model often places its centre
near the target but fails to size the box. This is the characteristic
tiny-target failure, where a 1--2\,px box error drops IoU below 0.5. These
numbers set a baseline that a future uncertainty-aware gating mechanism
would aim to improve upon.

\subsection{Qualitative Detection Across Stages}
\label{sec:vis}

\begin{figure*}[t]
  \centering
  \includegraphics[width=\textwidth]{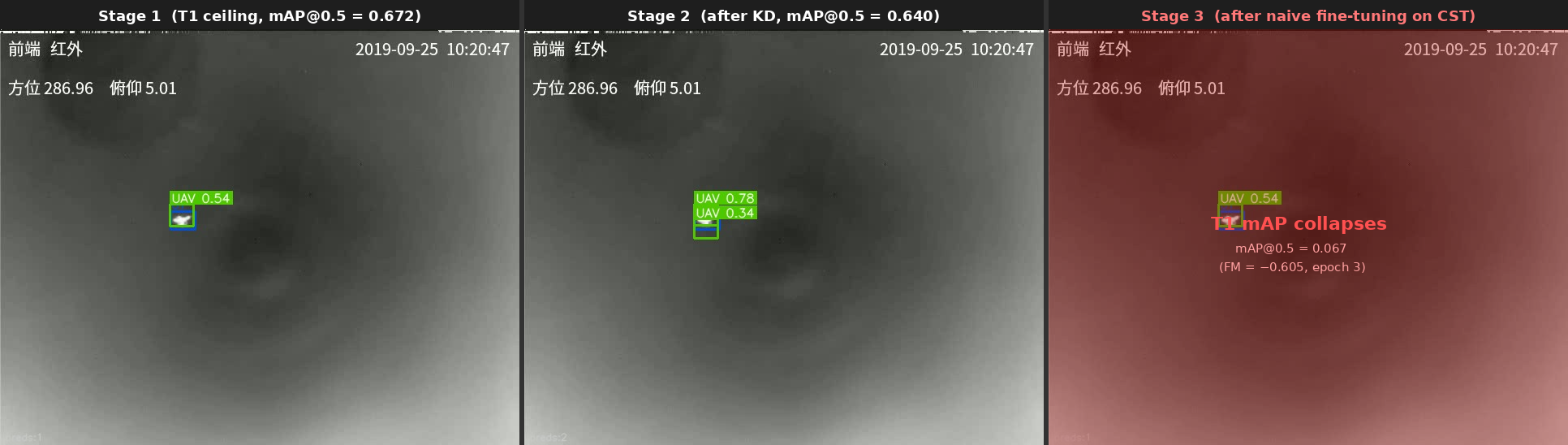}
  \caption{Detection output on the same Anti-UAV-RGBT validation frame
    (sequence 20190925\_101846, clip 1\_4, frame 0) under three successive
    checkpoints. Left: Stage 1 best checkpoint (T1 ceiling,
    mAP@0.5 = 0.6725). Centre: Stage 2 best checkpoint after knowledge
    distillation (mAP@0.5 = 0.640, FM = $-$0.033). Right: Stage 3 naive
    fine-tuning at epoch 3 (T1 mAP@0.5 = 0.068, FM = $-$0.605); the absence
    of large-target detections is in line with the gradient-starvation
    interpretation discussed above.}
  \label{fig:vis_comparison}
\end{figure*}

Figure~\ref{fig:vis_comparison} shows detection output on a single
Anti-UAV-RGBT validation frame under the three successive checkpoints. The
Stage 1 and Stage 2 models both localise the target confidently, whereas
the Stage 3 naive checkpoint produces no large-target detection, the
qualitative counterpart of the stratum-specific erasure quantified above.

\section{Discussion}
\label{sec:discussion}

\subsection{Comparison with Prior Continual Learning Results}

The Stage 2 Forgetting Measure of $-0.033 \pm 0.004$ is consistent with
what distillation-based continual learning achieves in general object
detection. Shmelkov et al.~\cite{shmelkov2017iod} introduced an
analogous distillation loss for incremental class detection on PASCAL VOC
and reported near-zero forgetting on old classes when the distillation
weight was tuned appropriately. Li and Hoiem~\cite{li2017lwf} applied a
similar output-distillation constraint for classification tasks and
likewise found that a frozen teacher constrains forgetting to within a
few percentage points under moderate domain shift. The 95\% T1 retention
observed here places KD in the same performance range. This offers early
evidence that output-level distillation transfers to the thermal anti-UAV
setting under a cross-dataset domain shift.

The Stage 3 result is harder to compare directly because most continual
learning benchmarks study class-incremental learning on balanced
datasets~\cite{wang2024comprehensive} rather than scale-distribution
shift on a single class. The closest analogue is the general finding that
forgetting is amplified when the new task distribution is maximally
dissimilar from the old one~\cite{Kirkpatrick_2017}. The 18-fold increase
from Stage 2 ($-0.033$) to Stage 3 ($-0.605$), with no change in the
protection mechanism, fits this principle. The Anti-UAV410-to-Anti-UAV-RGBT
gap is moderate (same modality, overlapping scales); the
CST-to-Anti-UAV-RGBT gap is extreme (near-complete scale inversion, zero
large-target overlap). What is novel is attributing this difference to a
specific measurable quantity, the scale-distribution shift, rather than a
qualitative description of task dissimilarity.

\subsection{Gradient Starvation as a Candidate Forgetting Mechanism}

The cosine similarity analysis in Table~\ref{tab:cosine_sim} produces a
counterintuitive result: Stage 3 moves the weights less than Stage 2 does,
relative to their starting point, yet Stage 3 forgetting is 18 times
larger. This holds whether cosine is measured over all parameters
(0.967 vs 0.911) or over the learnable, gradient-updated tensors alone
(0.987 vs 0.935). The latter is cleaner: it excludes the BatchNorm
running-statistic buffers, which are not updated by backpropagation. The standard account of catastrophic forgetting
attributes performance loss to large weight changes that overwrite prior
representations~\cite{mccloskey1989catastrophic,Kirkpatrick_2017}. Under
that account, smaller weight changes should produce less forgetting, not
more.

The gradient starvation framework~\cite{pezeshki2021gradient} offers a
plausible resolution to this contradiction. A representation that receives
little or no gradient signal during new-task training is not actively
overwritten by competing updates; the optimiser largely leaves it in
place. This is consistent with the measurements: the gradient-updated
weights move very little (learnable cosine 0.987), and the drift that does
occur is concentrated not in those weights but in the BatchNorm running
statistics (relative L2 $\approx$0.24 for
\texttt{running\_mean}/\texttt{running\_var} versus $\approx$0.12 for the
learnable weights). As the network re-estimates these normalisation
statistics for CST's tiny-target input distribution, the activations of
the now-unsupported large-target features can shift at test time even
though the underlying weights barely change. The outcome is a collapse
concentrated in exactly the strata absent from the new training
distribution: as Table~\ref{tab:stratum_forgetting} shows, large-stratum
mAP falls to near-zero within the first epoch and to 0.000 by epoch~1,
while the tiny stratum is essentially unaffected.

A direct gradient probe (Section~\ref{sec:gradprobe}) both supports and
qualifies this account. It indicates that gradient flow is
scale-conditioned: a head receives signal only in proportion to the
size-matched targets present, so a stratum absent from the data is left
unreinforced (the finest head is silent on T1, which has almost no tiny
($<$256\,px$^2$) targets, but strongly driven on tiny-dominated CST). It also shows,
however, that the detector's small anchors ($\leq 26$\,px) match
$\geq$64\,px targets only weakly, so the large-target pathway is fragile
even under T1, and that CST drives a box-regression loss roughly five
times larger than T1. The Stage~3 collapse is therefore most accurately
described as a \emph{scale-conditioned gradient imbalance} (the large
stratum receives no positive reinforcement while strong small-target
gradients dominate the shared heads): a refinement of, rather than a
departure from, the starvation interpretation.

This has a direct implication for method choice. Regularisation methods
such as EWC~\cite{Kirkpatrick_2017} penalise large weight changes, and
gradient-projection methods such as OGD~\cite{farajtabar2019ogd} and
GPM~\cite{saha2021gpm} project new-task updates orthogonal to old-task
directions; both target gradient \emph{interference}: new-task updates
overwriting old-task representations. The imbalance here is only partly of
that kind: the shared heads do receive strong small-target gradients
(Section~\ref{sec:gradprobe}), which a weight penalty could slow, but there
is no large-target gradient to project or preserve, since that signal is
absent from CST. Slowing the small-target updates cannot, on its own,
restore a large-target signal the new distribution never supplies. This
motivates a replay-based response that re-injects exemplars from the
under-represented strata, supplying the missing gradient directly, the
mechanism SSH provides.

\subsection{Limitations}

Several aspects of this study qualify the conclusions that can be drawn
from it.

The optimisation setup is not perfectly matched: Stage 3 used batch~16 and
learning rate $5\times10^{-4}$ against batch~64 and $10^{-3}$ for Stage 2
(Table~\ref{tab:apx:config}), so part of the larger Stage 3 forgetting could
reflect these settings rather than scale shift alone. Two facts limit this
concern: the forgetting is stratum-specific (a global change would not be
stratum-specific), and large-target mAP collapses within the first epoch,
before schedule
differences accumulate. A Stage 3 rerun at the Stage 2 settings would
isolate the effect.

Generalisability is bounded by the data: all three datasets are
predominantly DJI quadcopters in semi-controlled airspace, so the patterns
may not transfer to fixed-wing UAVs, evasive targets, or very different
backgrounds. The mechanism is also dataset-specific, depending on the
particular Anti-UAV410-to-CST scale inversion; a smaller scale gap might
shift the balance between weight drift and gradient imbalance.

On validity, the FM uses the Anti-UAV-RGBT validation split as its sole T1
reference, so a split biased toward easy sequences would understate
forgetting. Stage 2 is replicated across three seeds
(FM $= -0.033 \pm 0.004$), but Stage 3 is single-seed (seed 42), which
limits the precision of the Stage 3 estimate. We therefore claim a large,
qualitatively robust effect rather than a seed-precise value: the Stage 3
FM ($-0.605$) is more than two orders of magnitude larger than the Stage 2 seed-to-seed spread ($\pm 0.004$), the large-stratum
collapse is immediate (0.000 by epoch~1) and permanent, and T1 mAP decays
from 0.206 at epoch~0 to 0.017 at epoch~18 with no sustained recovery
(transient upticks $\leq$0.011)---none of which is characteristic
of a seed artefact, and all of
which is consistent with continual-evaluation work showing that forgetting
materialises within the first updates after a task
switch~\cite{delange2022stabilitygap}. Multi-seed replication of Stage 3
remains the first priority for future work.

Two design choices were not isolated. $\lambda_{\text{kd}}$ was fixed at 1.0
by convention rather than tuned (equal weighting follows
Li and Hoiem~\cite{li2017lwf} and Shmelkov et al.~\cite{shmelkov2017iod};
tuning it would need a held-out FM signal, hence a full S1$\to$S2$\to$S3 run
per value). And Stage 1 initialises from ImageNet weights, whose
large-object prior means the post-Stage-1 large-stratum capability may be
partly inherited.

A further limitation concerns causal attribution across stages. Stage~2 uses
knowledge distillation, whereas the headline Stage~3 baseline uses naive
fine-tuning, and the stages also differ in optimisation settings. Consequently,
the roughly 18-fold difference in FM magnitude cannot be attributed to scale
shift alone. In addition, because no Stage~2 no-KD baseline was run, the study
demonstrates strong retention under KD training but does not isolate how much
of that retention is caused by KD. A matched ablation with identical
optimisation settings and KD on/off at both transitions is required for a
causal comparison.

The replay comparison is single-seed and short. Random-stratified replay
slightly outperformed herding (FM $-0.221$ vs $-0.311$), suggesting the
scale stratification rather than the herding selection drives the benefit;
but with one seed and few epochs (five for SSH, three for the control) this
ordering between the two replay variants is indicative rather than
definitive. The buffer is drawn from the training split, disjoint from the
validation set used for the FM, so replay introduces no train--evaluation
overlap.

Finally, scalability was tested on a single architecture (YOLOMG); whether
the finding and SSH transfer to other YOLO variants or transformer
detectors is unknown, though the cosine diagnostic itself is
architecture-agnostic. Relatedly, all conclusions concern the learning
protocol rather than on-board deployment: training ran exclusively on HPC
infrastructure (Section~\ref{sec:compute}), and while inference-only use
of YOLO-class detectors on embedded hardware is plausible, the feasibility
of continual (re)training under edge constraints---an active research
area in its own right~\cite{pellegrini2021edge,park2025replay}---was
neither assumed nor evaluated.

\subsection{Value of the Research and Alternative Interpretations}

This research offers three contributions to the continual learning
literature. First, it sets out a concrete empirical protocol for measuring
catastrophic forgetting in thermal anti-UAV detection across real
operational datasets, producing the first quantified FM values for this
domain. Second, it puts forward a scale-conditioned gradient imbalance (a
form of gradient starvation) as a plausible mechanism under
scale-distribution shift, diagnosed with a lightweight inter-stage cosine
similarity that needs only the saved checkpoints. Third, it finds that
scale-stratified replay (a buffer balanced across size strata) roughly
halves the resulting forgetting. An ablation attributes this gain to the
\emph{stratification} (balanced coverage of size strata) rather than to the
iCaRL herding selection: the two stratified variants perform comparably,
whereas the no-replay baseline suffers the full collapse. This is
consistent with recent imbalanced
continual learning, where balancing the gradient contribution of
under-represented groups, not the specific exemplar-selection rule, is the
operative factor~\cite{he2024gradientreweighting,raghavan2024delta}. The
mechanism (well-represented scales degrading under a new
scale-imbalanced distribution) is not specific to anti-UAV detection: the
same imbalanced forgetting is reported in long-tailed continual
recognition~\cite{xu2024defying,molahasani2023continual}, so both the
diagnosis and the stratified-replay remedy should transfer to other scale-
or long-tailed vision settings. Whether the mechanism is the dominant one,
and whether the single-seed ordering between the replay variants holds,
cannot be fully settled from the present evidence.

An alternative interpretation of the Stage 3 result is that the
catastrophic collapse reflects not gradient starvation but a failure of
the Stage 2 checkpoint to generalise well to Anti-UAV-RGBT in the first
place: if the 95\% T1 retention at Stage 2 is fragile rather than
robust, even a small perturbation in Stage 3 could trigger a large FM
drop. This interpretation is not supported by the data. The per-stratum
breakdown (Table~\ref{tab:stratum_forgetting}) shows every T1 stratum
retained after Stage 2---the large stratum, which later collapses, is even
strengthened (0.461 to 0.494)---and the per-epoch tracking
(Table~\ref{tab:stage2_epochs}) shows stable T1 retention across all Stage
2 epochs. The collapse is Stage 3-specific and immediate, more consistent
with the scale-conditioned gradient imbalance than with pre-existing
fragility.

A second alternative is that the CST Anti-UAV annotations are noisier
than Anti-UAV-RGBT annotations for sub-10-pixel targets, and that the
detection loss instability from label noise rather than scale shift drives
the forgetting. This cannot be ruled out from the available evidence, but
the stratum-specific collapse pattern (large stratum affected most, tiny
stratum barely affected) is not what noisy label training would produce:
label noise would degrade all strata roughly uniformly, whereas gradient
starvation predicts exactly the stratum-specific pattern observed.

A third reading questions whether Stage 3 measures forgetting at all, since
the detector has almost no tiny-target capability before Stage 3. The two
are separated by construction: forgetting is claimed only for strata whose
capability was demonstrably acquired. The large stratum was
learned in Stage 1 (mAP 0.461) and further strengthened in Stage 2 (0.494)
before collapsing to 0.000 in Stage 3; this is forgetting of an acquired
capability, matching the ``similar-before, forget-after'' signature
documented for imbalanced continual learning~\cite{xu2024defying}. The tiny
stratum, by contrast, was never acquired (mAP $\approx$0 throughout), so no
forgetting is claimed for it; its low pre-Stage-3 value reflects a
plasticity limit of the small-anchor configuration rather than forgetting.

A note on YOLOMG's secondary input channel is warranted. Designed for
motion-difference masks, it is held at a zero tensor throughout:
motion masks were not generated for Anti-UAV410, CST Anti-UAV, or
Anti-UAV-RGBT within the scope of this study. Using Anti-UAV-RGBT's visible (RGB) stream as the Stage 1 secondary
channel was deliberately avoided, since Stages 2 and 3 are purely thermal:
the RGB channel would have to be zeroed mid-curriculum, adding a modality
shift that confounds the scale shift. Holding it at zero throughout keeps
the input modality constant, so the measured forgetting reflects scale
shift alone.

The research serves a cautionary, diagnostic role. Naive sequential
training across anti-UAV datasets appears unsuitable for deployment
(FM = $-0.605$ is a 90\% loss of T1 capability). Knowledge distillation
works as an effective first-line defence for moderate domain shift.
Scale-distribution shift, however, seems to call for a targeted
replay-based response that existing regularisation methods are not
designed to provide.

\section{Conclusion}
\label{sec:conclusion}

This research investigated catastrophic forgetting in a thermal
anti-UAV detector (YOLOMG, operated as a single thermal-infrared stream)
adapted sequentially across three heterogeneous
benchmarks: Anti-UAV-RGBT, Anti-UAV410, and CST Anti-UAV.

Under sequential training, YOLOMG exhibits measurable catastrophic
forgetting: naive fine-tuning on CST Anti-UAV produces a
Forgetting Measure of $-0.605$ against the Stage~1 ceiling, of which
$-0.572$ was incurred during Stage~3 alone, reducing Task 1 detection
capability to roughly 10\% of its post-Stage-1 level.

Stage~2 training with knowledge distillation yields
FM $= -0.033 \pm 0.004$ across three seeds, corresponding to 95\% T1
retention. This establishes that the KD-trained Stage~2 model preserves most
prior performance, but the absence of a Stage~2 no-KD control means the
causal benefit of distillation itself is not quantified here.

The Stage~3 results are consistent with scale-distribution shift acting as an amplifying factor. Moving
from the moderate domain gap of Stage 2 to the extreme scale inversion
of Stage 3 (CST Anti-UAV: 87.9\% tiny, 0\% large) coincides with an approximately 18-fold larger signed FM magnitude than the
Stage~2 transition under its KD training protocol. Large-target mAP collapses to near-zero within the first
Stage 3 epoch (0.000 by epoch~1) while the inter-stage cosine similarity over gradient-updated
weights remains high at 0.987 (0.967 over all parameters), higher than the
corresponding Stage 2 values (0.935 / 0.911). This dissociation argues
against weight drift as the primary mechanism. It points instead to a
scale-conditioned gradient imbalance, a refinement of gradient
starvation~\cite{pezeshki2021gradient}, as a candidate. The large stratum
receives no positive gradient signal from the new distribution, so its
(weakly anchored) output regime erodes despite minimal change in the
trained weights; the residual drift concentrates in BatchNorm running
statistics that track the new input. A direct gradient probe
(Section~\ref{sec:gradprobe}) is consistent with this scale-conditioned
signal flow.

Scale-Stratified Herding (SSH) targets this failure mode: a 300-exemplar
buffer selected by iCaRL-style greedy herding
within four UAV size strata (75 exemplars per stratum) re-injects the
missing gradient signal for underrepresented target scales. SSH roughly
halves the Forgetting Measure ($-0.605 \to -0.311$) and keeps large-target
T1 mAP non-zero (0.079 vs 0.000), at a small cost to new-task detection. A
random-stratified control performs at least as well ($-0.221$), indicating
the gain comes from the scale stratification rather than the herding selection.

These conclusions are qualified by several limitations. Stage 3 was
trained at a smaller batch size and lower learning rate than Stage 2, so
part of the measured forgetting may reflect optimisation differences as
well as scale shift. The gradient-starvation finding also depends on the
particular scale inversion between Anti-UAV410 and CST Anti-UAV and may
not generalise to dataset pairs with smaller scale gaps. Finally, the
Stage 3 and replay results rest on a single seed and short (few-epoch)
runs, so the small margin between the two replay variants is indicative
rather than definitive.

The primary contribution of this research is empirical: it provides a
quantified forgetting protocol for thermal anti-UAV detection, and
suggests that naive sequential training across operational datasets is
unsuitable for deployed C-UAV systems (FM $= -0.605$, 90\% capability
loss). Three directions for future work follow from these
findings. Most immediately, the single-seed, few-epoch replay results
should be replicated across multiple seeds and longer schedules to confirm
the effect size and the currently narrow margin between herding and random
replay. Comparing the stratified-replay response against regularisation
methods such as EWC~\cite{Kirkpatrick_2017} and
GEM~\cite{lopezpaz2022gradientepisodicmemorycontinual} would clarify whether
replay or regularisation better counters the imbalance under scale shift.
Finally, integrating the detector with uncertainty-aware
tracking~\cite{zhu_evidential_2023,sensoy2018evidential}, improving on the
tracking baseline of Section~\ref{sec:tracking}, would extend the
framework to the full detection-to-tracking pipeline, addressing the
complete C-UAV operational scenario.

\bibliographystyle{ACM-Reference-Format}
\bibliography{bibtex_acm_clean}

\appendix

\section{Dataset Statistics}
\label{sec:apx:datasets}

\begin{figure}[h]
  \centering
  \includegraphics[width=\linewidth]{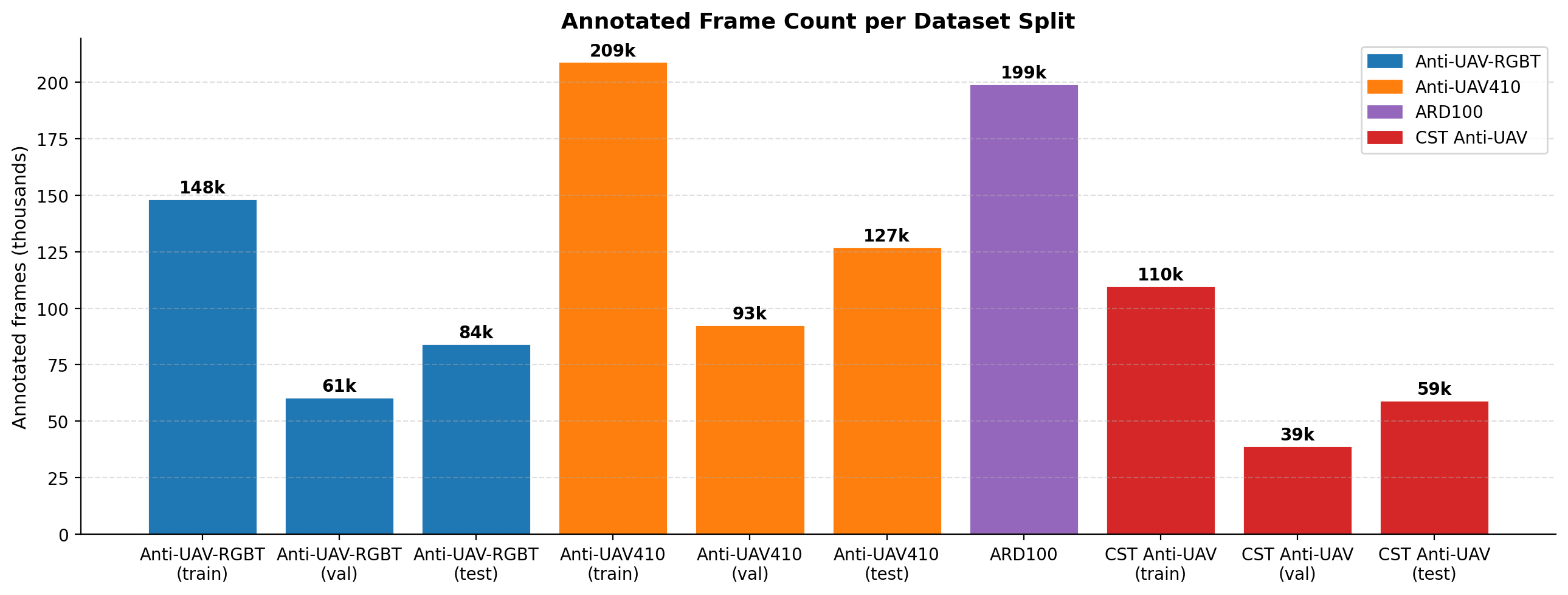}
  \caption{Per-split counts of UAV-present frames---frames in which a UAV is
    present (the \texttt{exist}=1 flag), labelled `annotated' on the figure
    axis. Anti-UAV410 is the largest (428,703 across train/val/test),
    followed by Anti-UAV-RGBT (293,209). CST Anti-UAV (208,221) and ARD100
    (199,291) are comparable in size; ARD100 is used only for motion-mask
    pre-computation and is not part of the continual learning curriculum.}
  \label{fig:apx:frame_counts}
\end{figure}

\begin{figure}[h]
  \centering
  \includegraphics[width=\linewidth]{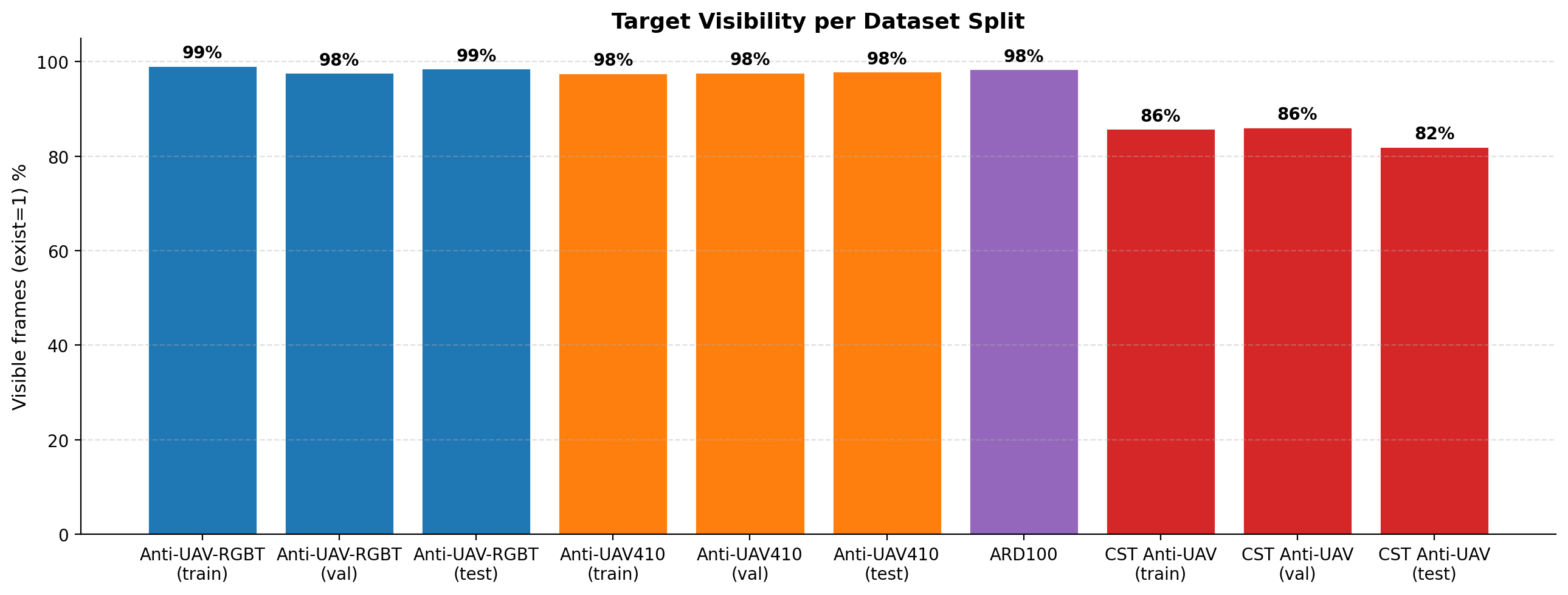}
  \caption{Target visibility per dataset split. A frame is marked
    visible if it contains an actively flying UAV (the \texttt{exist}=1 flag);
    occluded or out-of-frame instances are marked absent. Anti-UAV-RGBT
    has the highest visibility rate, CST Anti-UAV the lowest, consistent
    with the difficulty of detecting sub-16-pixel targets in cluttered
    thermal scenes.}
  \label{fig:apx:visibility}
\end{figure}

Figure~\ref{fig:apx:frame_counts} reports the per-split UAV-present frame
counts for each dataset, and Figure~\ref{fig:apx:visibility} the
corresponding target-visibility rates.

\section{Experimental Configuration}
\label{sec:apx:config}

Table~\ref{tab:apx:config} summarises the hyperparameters used across all
three training stages. All stages use Nesterov SGD with cosine learning
rate decay and the same YOLOMG architecture (318 layers, 335 named
parameter tensors, single UAV class).

\begin{table*}[t]
  \centering
  \caption{Training configuration per stage. Stage 2 is run with three
  seeds; Stages 1 and 3 use seed 42 only. Epochs are 0-indexed
  (epoch~0 is the first).}
  \label{tab:apx:config}
  \small

  \begin{tabularx}{\textwidth}{
      >{\raggedright\arraybackslash}p{0.15\textwidth}
      >{\raggedright\arraybackslash}X
      >{\raggedright\arraybackslash}X
      >{\raggedright\arraybackslash}X
    }
    \toprule
    & \textbf{Stage 1}
    & \textbf{Stage 2}
    & \textbf{Stage 3} \\
    \midrule

    Dataset
      & Anti-UAV-RGBT (IR)
      & Anti-UAV410
      & CST Anti-UAV \\

    Objective
      & Supervised det.
      & KD + det.
      & Naive / replay det.$^\dagger$ \\

    Epochs
      & 100 (early stop)
      & 50 (early stop)
      & Naive: 19 epochs, best ep.~3;
        SSH: 5 epochs, best ep.~2;
        random: 3 epochs, best ep.~2 \\

    Best checkpoint
      & Epoch 49
      & Ep.~16/15/27 (seeds 42/123/999)
      & See Epochs row \\

    Optimiser
      & SGD (Nesterov)
      & SGD (Nesterov)
      & SGD (Nesterov) \\

    Initial LR
      & $10^{-2}$
      & $10^{-3}$
      & $5\times10^{-4}$ \\

    LR schedule
      & Cosine decay
      & Cosine decay
      & Cosine decay \\

    Weight decay
      & $5\times10^{-4}$
      & $5\times10^{-4}$
      & $5\times10^{-4}$ \\

    Momentum
      & 0.937
      & 0.937
      & 0.937 \\

    Batch size
      & 64 (4$\times$A100)
      & 64 (4$\times$H100)
      & 16 (4$\times$A100) \\

    $\lambda_{\mathrm{kd}}$
      & ---
      & 1.0
      & --- \\

    Seeds
      & 42
      & 42, 123, 999
      & 42 \\

    Input ch.\ 1
      & IR frame
      & IR frame
      & IR frame \\

    Input ch.\ 2
      & Zeros
      & Zeros
      & Zeros \\

    \bottomrule
  \end{tabularx}

  \vspace{2pt}
  \begin{minipage}{\textwidth}
    \footnotesize
    $^\dagger$Three Stage~3 variants were run: naive (no replay),
    SSH (scale-stratified herding replay), and random-stratified replay.
    All share the same hyperparameters; replay variants additionally use
    replay weight $=4.0$, replay ratio $=25\%$, and buffer size $=300$
    (75 per stratum).
  \end{minipage}
\end{table*}

\section{Additional Training Dynamics}
\label{sec:apx:dynamics}

Table~\ref{tab:stage2_epochs} reports the epoch-level Stage 2 dynamics
(seed 42) and Table~\ref{tab:s3_naive_epochs} the per-epoch Stage 3 naive
dynamics; both are discussed in the main text
(Sections~\ref{sec:rq1} and~\ref{sec:h1}).

\begin{table}[h]
  \centering
  \caption{Epoch-level Stage 2 results (seed 42, 32 epochs).
    kd/det is the ratio of $L_{\text{kd}}$ to $L_{\text{det}}$,
    averaged over batches. Epoch 16 is the peak T2 checkpoint. FM values
    here are seed~42's Stage~2 training-time evaluation (e.g. $-0.037$ at
    epoch~16). The post-Stage-2 baseline used for the headline Forgetting
    Measure and as the Stage~3 starting point is a clean re-evaluation of
    this checkpoint (T1 mAP $0.640$, FM $-0.033$); the $\sim$0.004 gap
    reflects training-time versus re-evaluation differences. The three-seed
    mean Stage~2 FM is $-0.033 \pm 0.004$.}
  \label{tab:stage2_epochs}
  \small
  \begin{tabular}{@{}crrrrr@{}}
    \toprule
    Epoch & $L_{\text{det}}$ & $L_{\text{kd}}$ & kd/det & T2 mAP & FM \\
    \midrule
    0  & 0.0580 & 0.0791 & 1.36 & 0.403 & $-$0.007 \\
    5  & 0.0355 & 0.0768 & 2.16 & 0.419 & $-$0.035 \\
    10 & 0.0317 & 0.0739 & 2.33 & 0.406 & $-$0.046 \\
    15 & 0.0293 & 0.0712 & 2.43 & 0.417 & $-$0.042 \\
    16 & 0.0289 & 0.0705 & 2.44 & 0.426 & $-$0.037 \\
    20 & 0.0273 & 0.0677 & 2.48 & 0.419 & $-$0.041 \\
    25 & 0.0256 & 0.0648 & 2.54 & 0.415 & $-$0.044 \\
    31 & 0.0233 & 0.0612 & 2.62 & 0.422 & $-$0.037 \\
    \bottomrule
  \end{tabular}
\end{table}

\begin{table}[h]
  \centering
  \caption{Stage 3 naive condition — per-epoch results. Pre-training T1
    mAP (before any Stage~3 gradient steps) = 0.640. T1 ceiling = 0.6725.
    Bold marks the best T3 checkpoint (epoch~3). Large-stratum T1 mAP
    reaches 0.000 at epoch~1 and never recovers; the small and normal
    strata fall steeply in parallel (from 0.559 and 0.684 post-Stage-2,
    Table~\ref{tab:stratum_forgetting}, to 0.060 and 0.079 at the best
    checkpoint).}
  \label{tab:s3_naive_epochs}
  \scriptsize
  \setlength{\tabcolsep}{2.2pt}
  \begin{tabular}{@{}crrrrrrrr@{}}
    \toprule
    Ep. & T3 mAP & T1 mAP & FM$_{\text{abs}}$ & FM$_{\text{stage3}}$
        & Tiny & Small & Normal & Large \\
    \midrule
    pre & ---   & 0.640 & $-$0.033 &  $\phantom{-}$0.000 & ---   & ---   & ---   & ---   \\
    0   & 0.042 & 0.206 & $-$0.466 & $-$0.434 & 0.041 & 0.130 & 0.251 & 0.010 \\
    1   & 0.066 & 0.127 & $-$0.545 & $-$0.513 & 0.005 & 0.090 & 0.151 & 0.000 \\
    2   & 0.074 & 0.082 & $-$0.591 & $-$0.559 & 0.001 & 0.075 & 0.092 & 0.000 \\
    \textbf{3} & \textbf{0.083} & \textbf{0.068} & $-$\textbf{0.605} & $-$\textbf{0.572} & \textbf{0.001} & \textbf{0.060} & \textbf{0.079} & \textbf{0.000} \\
    4   & 0.066 & 0.057 & $-$0.615 & $-$0.583 & 0.000 & 0.056 & 0.065 & 0.000 \\
    5   & 0.071 & 0.053 & $-$0.620 & $-$0.588 & 0.000 & 0.042 & 0.064 & 0.000 \\
    6   & 0.063 & 0.048 & $-$0.624 & $-$0.592 & 0.000 & 0.052 & 0.055 & 0.000 \\
    7   & 0.076 & 0.039 & $-$0.633 & $-$0.601 & 0.000 & 0.045 & 0.044 & 0.000 \\
    18  & 0.062 & 0.017 & $-$0.656 & $-$0.623 & 0.000 & 0.019 & 0.019 & 0.000 \\
    \bottomrule
  \end{tabular}
\end{table}

\section{Preprocessing Pipelines}
\label{sec:apx:preprocessing}

\subsection*{Motion Mask Generation (mask32)}

YOLOMG's second input channel is designed to carry a pixel-level motion
difference map (mask32), computed from five consecutive frames after global
motion compensation (GMC). GMC uses the Enhanced Correlation Coefficient
(ECC) algorithm~\cite{evangelidis2008parametric} to estimate the
camera's ego-motion between frames, warp earlier frames to align with the
current frame, and then compute the frame-difference residual. The residual
retains only object-relative motion, suppressing camera shake.

The five-frame sliding window operates as follows. Let frames be indexed
$t-2, t-1, t, t+1, t+2$. The FD5 scheme uses three non-adjacent frames
($t-2$, $t$, and $t+2$) to compute the motion difference after ECC-based
alignment. For Stage 1 (Anti-UAV-RGBT) and all purely thermal datasets
(Anti-UAV410, CST Anti-UAV), motion masks were not available within the
scope of this study, so the second input channel is set to a zero tensor
throughout. This means the motion branch of YOLOMG contributes no signal
during training and the model functions as a single-stream thermal
appearance detector.

\subsection*{Dataset Loading Strategy}

To avoid inode quota exhaustion on Snellius, no intermediate converted files
are written to disk. Each dataset is read in its
native format inside a custom PyTorch \texttt{Dataset} class.
Table~\ref{tab:apx:loading} summarises the loading strategy per dataset.

\begin{table}[h]
  \centering
  \caption{Native-format loading strategy per dataset. No intermediate
    converted files are written; format-specific logic is handled inside the
    custom \texttt{Dataset} implementation.}
  \label{tab:apx:loading}
  \scriptsize
  \begin{tabularx}{\columnwidth}{@{}l>{\raggedright\arraybackslash}X>{\raggedright\arraybackslash}X@{}}
    \toprule
    Dataset & Frame source & Annotation source \\
    \midrule
    Anti-UAV-RGBT & MP4 video (OpenCV) & \texttt{infrared.json} \\
    Anti-UAV410 & Pre-extracted JPEG & \texttt{IR\_label.json} \\
    CST Anti-UAV & Pre-extracted JPEG & \texttt{gt.txt} + \texttt{IR\_label.json} \\
    ARD100 & MP4 video (OpenCV) & Pascal VOC XML per frame \\
    \bottomrule
  \end{tabularx}
\end{table}

\subsection*{Frame Extraction}

Anti-UAV-RGBT stores sequences as \texttt{.mp4} video files with
per-frame JSON annotations. Frames are decoded on demand using
\texttt{cv2.VideoCapture} with frame-index seeking (or pre-extracted as
JPEG files to eliminate random-seek overhead on NFS mounts). Only the
infrared stream (\texttt{infrared.mp4}) is loaded; the paired RGB stream
(\texttt{visible.mp4}) is ignored entirely. Anti-UAV410 and CST Anti-UAV
store pre-extracted JPEG frames, which are read directly.

\subsection*{Scale Stratification}

Bounding box scale stratification is applied consistently across all
scripts using bounding-box area in pixels. A normalised box $(w_n, h_n)$
is first converted to pixel dimensions at the canonical resolution
$640 \times 512$:
\[
  a = (w_n \times 640)\times(h_n \times 512)
\]
The four strata are:
\begin{itemize}
  \item tiny: $a < 256\,\text{px}^2$ ($= 16^2$)
  \item small: $256 \leq a < 1024\,\text{px}^2$ ($16^2$--$32^2$)
  \item normal: $1024 \leq a < 4096\,\text{px}^2$ ($32^2$--$64^2$)
  \item large: $a \geq 4096\,\text{px}^2$ ($\geq 64^2$)
\end{itemize}
The same thresholds are used consistently for evaluation, scale-distribution
analysis, and replay-buffer construction.

\section{Pseudocode}
\label{sec:apx:pseudocode}

\subsection*{Algorithm 1: Scale-Stratified Herding (SSH)}

Algorithm~\ref{alg:ssh} formalises the SSH buffer construction. The key
departure from standard iCaRL herding is that the T1 reference data are
first partitioned by target-size stratum before greedy herding is applied,
ensuring representation of all four scale categories regardless of their
frequency in the source dataset.

\begin{algorithm}[t]
\caption{Scale-Stratified Herding (SSH)}
\label{alg:ssh}
\begin{algorithmic}[1]

\Require T1 reference set $\mathcal{D}_1$,
Stage~2 encoder $f_{\theta_2}$,
exemplars per stratum $K=75$
\Ensure Exemplar buffer $\mathcal{M}$ with $|\mathcal{M}|=300$

\State Define
$\mathcal{S}\gets
\{\text{tiny},\text{small},\text{normal},\text{large}\}$

\For{each $x\in\mathcal{D}_1$}
    \State $e(x)\gets
    \mathrm{GlobalAvgPool}\!\left(f_{\theta_2}^{P3}(x)\right)$
    \State Assign size stratum $s(x)$ from bounding-box area
\EndFor

\State $\mathcal{M}\gets\emptyset$

\For{each $s\in\mathcal{S}$}

    \State $\mathcal{D}_s
    \gets
    \{x\in\mathcal{D}_1:s(x)=s\}$

    \State $\displaystyle
    \mu_s\gets
    \frac{1}{|\mathcal{D}_s|}
    \sum_{x\in\mathcal{D}_s}e(x)$

    \State $\text{selected}\gets[\,]$
    \State $\bar{e}\gets\mathbf{0}$

    \For{$k=1,\ldots,K$}

        \State $\displaystyle
        j^*\gets
        \arg\min_{j\notin\text{selected}}
        \left\|
        \frac{(k-1)\bar{e}+e(x_j)}{k}
        -\mu_s
        \right\|_2$

        \State $\text{selected}.\mathrm{append}(j^*)$

        \State $\displaystyle
        \bar{e}\gets
        \frac{(k-1)\bar{e}+e(x_{j^*})}{k}$

    \EndFor

    \State $\mathcal{M}\gets
    \mathcal{M}\cup
    \{x_j:j\in\text{selected}\}$

\EndFor

\State \Return $\mathcal{M}$

\end{algorithmic}
\end{algorithm}

\subsection*{Algorithm 2: Stage 2 Knowledge Distillation}

Algorithm~\ref{alg:kd} summarises the Stage~2 training procedure. The
teacher is initialised from the Stage~1 checkpoint and remains frozen
throughout training. The student and teacher receive the same infrared
input, while the student is additionally supervised using the
Anti-UAV410 ground-truth annotations.

\begin{algorithm}[t]
\caption{Stage 2 Knowledge Distillation Training}
\label{alg:kd}
\begin{algorithmic}[1]

\Require Stage~1 checkpoint $\theta_T$,
training set $\mathcal{D}_2$,
$\lambda_{\mathrm{kd}}=1.0$,
epochs $E$
\Ensure Stage~2 student checkpoint $\theta_S^*$

\State $\theta_S\gets\theta_T$
\State Freeze $\theta_T$
\State Set $f_{\theta_T}$ to evaluation mode

\For{$e=1,\ldots,E$}

    \For{each mini-batch $(x,y)\in\mathcal{D}_2$}

        \State $\mathbf{p}_S
        \gets f_{\theta_S}(x,\mathbf{0})$

        \State $\mathbf{p}_T
        \gets f_{\theta_T}(x,\mathbf{0})$

        \State $\mathcal{L}_{\mathrm{det}}
        \gets
        \mathrm{YOLOLoss}(\mathbf{p}_S,y)$

        \State $\displaystyle
        \mathcal{L}_{\mathrm{kd}}
        \gets
        \frac{1}{3}
        \sum_{p\in\{P3,P4,P5\}}
        \mathrm{MSE}
        \left(
        \mathbf{p}_S^{(p)},
        \mathbf{p}_T^{(p)}
        \right)$

        \State $\mathcal{L}
        \gets
        \mathcal{L}_{\mathrm{det}}
        +
        \lambda_{\mathrm{kd}}
        \mathcal{L}_{\mathrm{kd}}$

        \State Update $\theta_S$ using SGD on
        $\nabla_{\theta_S}\mathcal{L}$

    \EndFor

    \State Evaluate $\theta_S$ on T1 and T2 validation sets

\EndFor

\State $\displaystyle
\theta_S^*
\gets
\arg\max_e
\mathrm{mAP@0.5}_{T2}
\left(\theta_S^{(e)}\right)$

\State \Return $\theta_S^*$

\end{algorithmic}
\end{algorithm}

\subsection*{Algorithm 3: Forgetting Measure Computation}

Algorithm~\ref{alg:fm} defines the Forgetting Measure used throughout this
research. The T1 performance obtained immediately after Stage~1 is fixed
as the reference baseline against which retention after subsequent
training stages is measured.

\begin{algorithm}[t]
\caption{Forgetting Measure (FM)}
\label{alg:fm}
\begin{algorithmic}[1]

\Require Stage~1 checkpoint $\theta_1$,
Stage~$k$ checkpoint $\theta_k$,
T1 validation set $\mathcal{D}_1^{\mathrm{val}}$
\Ensure Forgetting Measure $\mathrm{FM}$

\State $m_1\gets
\mathrm{mAP@0.5}
\left(
f_{\theta_1},
\mathcal{D}_1^{\mathrm{val}}
\right)$

\State $m_k\gets
\mathrm{mAP@0.5}
\left(
f_{\theta_k},
\mathcal{D}_1^{\mathrm{val}}
\right)$

\State $\mathrm{FM}\gets m_k-m_1$

\State \Return $\mathrm{FM}$

\end{algorithmic}
\end{algorithm}

\section{Stage 3 Tracking Evaluation}
\label{sec:apx:tracking}

\begin{figure}[ht]
  \centering
  \includegraphics[width=0.8\linewidth]{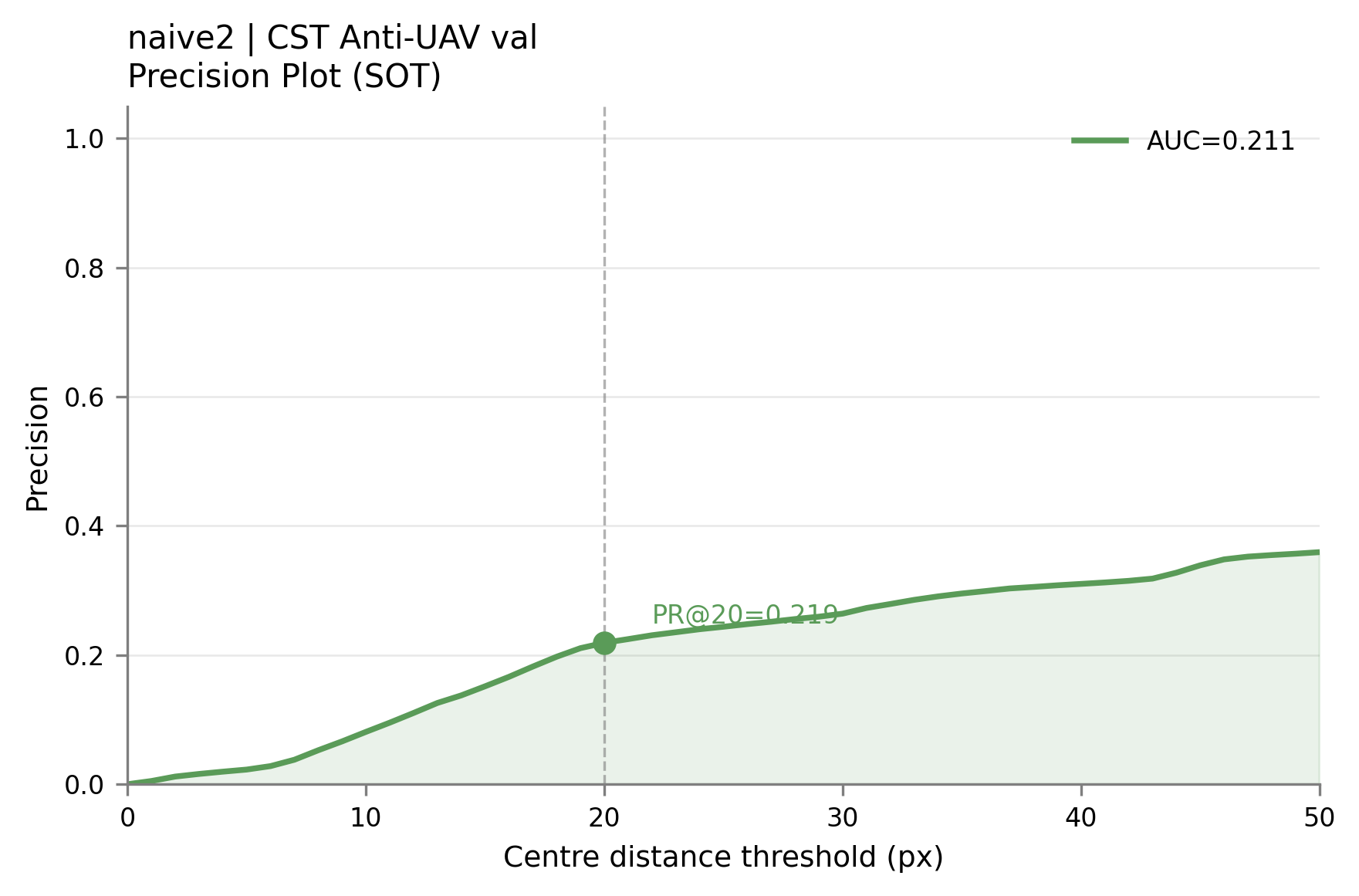}
  \caption{Precision plot on CST Anti-UAV validation set (Stage 3 best
    checkpoint, epoch 3). The representative score at threshold 20\,px
    (PR@20) is reported in the main tracking evaluation table.}
  \label{fig:apx:precision}
\end{figure}

\begin{figure}[ht]
  \centering
  \includegraphics[width=0.8\linewidth]{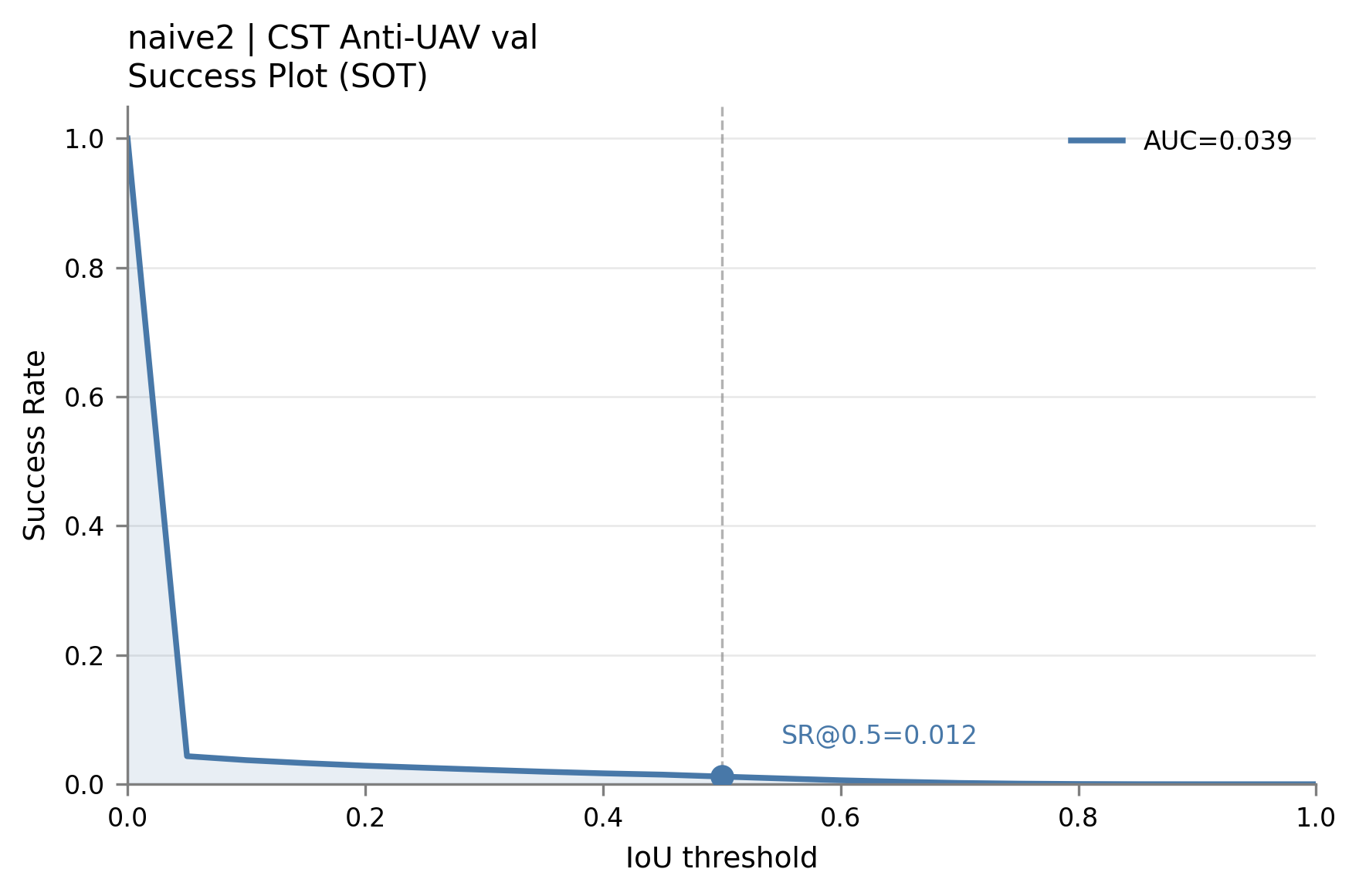}
  \caption{Success plot on CST Anti-UAV validation set (Stage 3 best checkpoint,
    epoch 3). The area under this curve (AUC) provides an
    IoU-threshold-independent summary of tracking quality.}
  \label{fig:apx:success}
\end{figure}

\begin{figure}[ht]
  \centering
  \includegraphics[width=\linewidth]{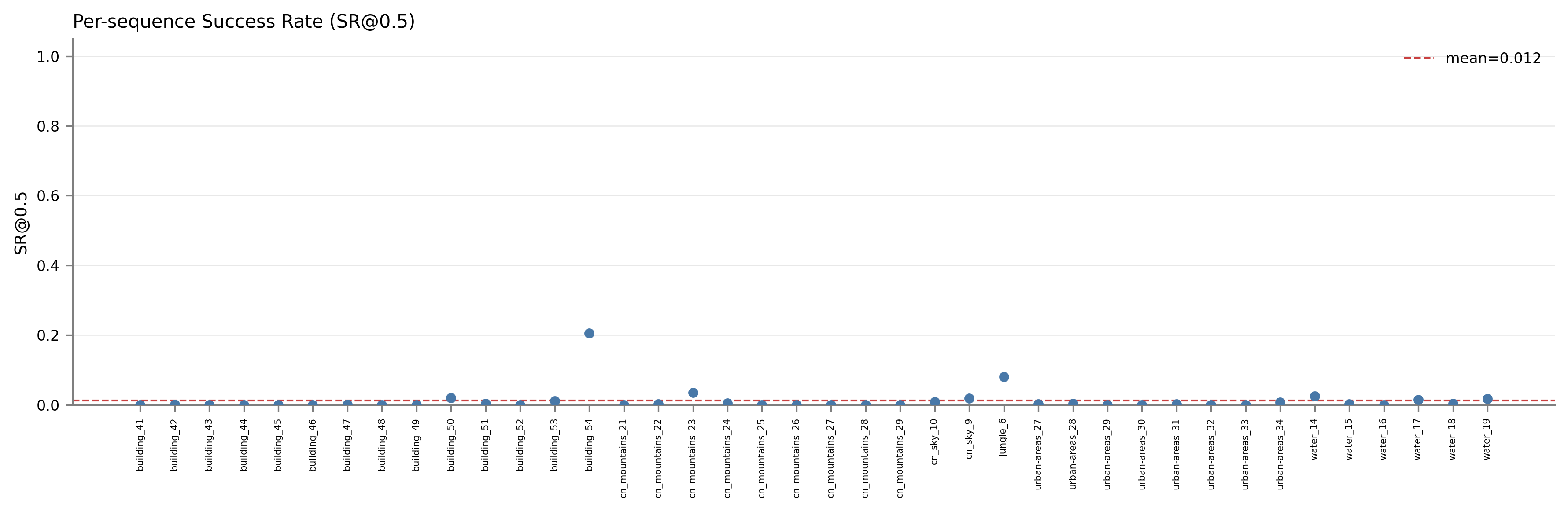}
  \caption{Per-sequence success rate at IoU~0.5 (SR@0.5) on CST Anti-UAV
    validation set. Sequences with predominantly tiny targets ($<$256\,px$^2$) show
    the lowest SR@0.5, consistent with the near-zero tiny-stratum mAP
    reported in Table~\ref{tab:stratum_forgetting}.}
  \label{fig:apx:seq_sr}
\end{figure}

Figures~\ref{fig:apx:precision}--\ref{fig:apx:seq_sr} report the
one-pass evaluation (OPE) tracking results for the Stage 3 best checkpoint
on the CST Anti-UAV validation set, using the protocol of
Huang et al.~\cite{bohuang_410}. The precision plot
(Figure~\ref{fig:apx:precision}) shows the proportion of frames whose
predicted centre falls within a threshold pixel distance of the ground-truth
centre; the success plot (Figure~\ref{fig:apx:success}) shows the
proportion of frames with IoU overlap above threshold.
Figure~\ref{fig:apx:seq_sr} breaks success rate at IoU 0.5 (SR@0.5)
down by sequence, revealing the substantial per-sequence variance that is
not visible in the aggregate mAP.

\clearpage
\section{Compute Environment}
\label{sec:apx:env}

All training and analysis jobs were executed on the Snellius HPC cluster
(SURF, the Netherlands). The software environment was a dedicated Conda
environment (Python~3.9) created with Conda. Packages were installed with \texttt{pip}
inside the environment. User-specific filesystem paths are omitted from this
public preprint.

\subsubsection*{Library versions}

\begin{center}\small
\begin{tabular}{ll}
\hline
\textbf{Package} & \textbf{Version} \\
\hline
Python           & 3.9 \\
torch            & 2.7.1+cu118 \\
torchvision      & 0.22.1+cu118 \\
torchaudio       & 2.7.1+cu118 \\
numpy            & 2.0.2 \\
\hline
\end{tabular}
\end{center}

\subsubsection*{Environment activation}

Experiments were run in a dedicated Python~3.9 Conda environment after
loading the Snellius 2023 software stack and Miniconda module. User-specific
filesystem paths are intentionally omitted from this public preprint. Packages
without a recorded pinned version are not listed above.

\section{Experimental Run Summary}
\label{sec:apx:slurm}

Table~\ref{tab:apx:slurm} summarises the runs underlying the reported
results. Internal scheduler job identifiers are omitted from the public
preprint because they are cluster-specific and do not improve reproducibility.

\begin{table}[h]
  \centering
  \caption{Experimental runs underlying the reported results.}
  \label{tab:apx:slurm}
  \small
  \begin{tabular}{@{}llp{0.50\linewidth}@{}}
    \toprule
    Run & Seed(s) & Notes \\
    \midrule
    Stage 1 supervised & 42 & Best mAP@0.5 = 0.6725 at epoch 49 \\
    Stage 2 KD & 42, 123, 999 & Three-seed retention analysis \\
    Stage 3 naive & 42 & FM = $-$0.605; best T3 checkpoint at epoch 3 \\
    Stage 3 SSH replay & 42 & FM = $-$0.311; best T3 checkpoint at epoch 2 \\
    Stage 3 random replay & 42 & FM = $-$0.221; best T3 checkpoint at epoch 2 \\
    Gradient probe & 42, 123, 999 & Stage~2 $\to$ Stage~3 boundary analysis \\
    \bottomrule
  \end{tabular}
\end{table}

\clearpage

\end{document}